\documentclass{article}
\usepackage{geometry}

\usepackage[utf8]{inputenc} 
\usepackage[T1]{fontenc}    
\usepackage[pagebackref=true]{hyperref}       
\usepackage{url}            
\usepackage{booktabs}       
\usepackage{amsfonts}       
\usepackage{nicefrac}       
\usepackage{microtype}      
\usepackage{xcolor}         
\usepackage{amsmath}
\usepackage{algorithm}
\usepackage{algorithmic}
\usepackage{bbm}
\usepackage{subfig}
\usepackage{graphicx}
\usepackage{mathtools}
\usepackage{pdflscape}
\usepackage{listings}
\usepackage{caption}
\usepackage{amsthm}
\usepackage{enumitem}
\usepackage{authblk}
\usepackage[round]{natbib}
\usepackage{mathrsfs}

\renewcommand*\backref[1]{\ifx#1\relax \else (Cited on #1) \fi}

\newcommand{\E}{\mathbb{E}}

\newcommand{\N}{\mathbb{N}}

\newcommand{\R}{\mathbb{R}}

\newcommand{\cD}{\mathcal{D}}

\newcommand{\cH}{\mathcal{H}}

\newcommand{\cL}{\mathcal{L}}

\newcommand{\cO}{\mathcal{O}}

\newcommand{\cR}{\mathcal{R}}

\newcommand{\cX}{\mathcal{X}}
\newcommand{\cY}{\mathcal{Y}}
\newcommand{\cZ}{\mathcal{Z}}

\newcommand{\bfC}{\mathbf{C}}

\newcommand{\bfL}{\mathbf{L}}
\newcommand{\bfM}{\mathbf{M}}

\newcommand{\bfP}{\mathbf{P}}
\newcommand{\bfQ}{\mathbf{Q}}

\newcommand{\bfZ}{\mathbf{Z}}

\newcommand{\bfLambda}{\boldsymbol{\Lambda}}
\newcommand{\bfPsi}{\boldsymbol{\Psi}}

\theoremstyle{plain}
\newtheorem{theorem}{Theorem}[section]

\newtheorem{lemma}[theorem]{Lemma}
\newtheorem{corollary}[theorem]{Corollary}
\theoremstyle{definition}
\newtheorem{definition}[theorem]{Definition}
\newtheorem{assumption}[theorem]{Assumption}

\theoremstyle{remark}
\newtheorem{remark}[theorem]{Remark}

\DeclareMathOperator*{\argmin}{arg\,min}

\newcommand{\ip}[2] {\langle #1, #2 \rangle }

\newcommand{\tr}{\mathrm{trace}}

\newcommand{\var}{\text{Var}}

\newcommand{\Id}{\mathbf{Id}}
\newcommand{\UCB}{\mathrm{UCB}}
\newcommand{\LCB}{\mathrm{LCB}}

\title{Random Exploration in Bayesian Optimization: Order-Optimal Regret and Computational Efficiency}
\author[*]{Sudeep Salgia}
\author[$\ddagger$]{Sattar Vakili}
\author[*]{Qing Zhao}
\affil[*]{School of Electrical \& Computer Engineering, Cornell University, Ithaca, NY, \emph{\{ss3827,qz16\}@cornell.edu} }
\affil[$\ddagger$]{MediaTek Research, UK, \emph{sattar.vakili@mtkresearch.com}}
\date{Oct 2023; Revised Feb 2024}

\begin{document}

\maketitle

\begin{abstract}
    We consider Bayesian optimization using Gaussian Process models, also referred to as kernel-based bandit optimization. We study the methodology of exploring the domain using random samples drawn from a distribution. We show that this random exploration approach achieves the optimal error rates. Our analysis is based on novel concentration bounds in an infinite dimensional Hilbert space established in this work, which may be of independent interest. We further develop an algorithm based on random exploration with domain shrinking and establish its order-optimal regret guarantees under both noise-free and noisy settings. In the noise-free setting, our analysis closes the existing gap in regret performance and thereby \emph{resolves a COLT open problem}. The proposed algorithm also enjoys a computational advantage over prevailing methods due to the random exploration that obviates the expensive optimization of a non-convex acquisition function for choosing the query points at each iteration.                
\end{abstract}

\section{Introduction}
\label{sec:introduction}

\subsection{GP-based Bayesian Optimization}

We consider the problem of sequential optimization of an unknown, possibly non-convex, function $f: \cX \to \R$. The learner sequentially chooses a query point $x_t \in \cX$ at each time $t$ and observes the function value (potentially subject to noise) at $x_t$. The learning objective is to approach a global maximizer $x^*$ of the function through a sequence of query points $\{x_t\}_{t = 1}^T$ chosen sequentially in time. In addition to the convergence of $\{x_t\}_{t = 1}^T$ to $x^*$, an online measure of the learning efficiency is the \emph{cumulative regret}
\begin{align} 
\label{eqn:regret_definition}
	R(T) = \sum_{t = 1}^T \left[ f(x^*) - f(x_t) \right]. 
\end{align}

The above problem finds a wide range of applications including hyperparameter optimization~\cite{Li2016Hyperband}, experimental design~\cite{Greenhill2020Bayesian}, recommendation systems~\cite{Vanchinathan2014Explore} and robotics~\cite{Lizotte2007Automatic}. An approach that has proven to be particularly effective is Bayesian Optimization (BO) using Gaussian Process (GP) models (a.k.a. kernel-based bandit optimization). The unknown objective function $f$ is assumed to live in a Reproducing Kernel Hilbert Space (RKHS) associated with a known kernel. Within the GP-based BO framework, $f$ is viewed as a realization of a Gaussian process over $\cX$. With each new query $x_t$, the learner sharpens the posterior distribution and uses it as a proxy for $f$ for subsequent optimization. We point out that such a Bayesian approach is equally applicable to a frequentist formulation where $f$ is \emph{deterministic} as considered in this work. In this case, the GP model of $f$ is fictitious and internal to the algorithm. \\

Under the assumption of noise-free query feedback, BO techniques were used for optimization as early as 1964~\cite{Kushner1964}. GP-based BO was popularized through the work of~\citet{Mockus1978}. Since then, a number of approaches have been developed and analyzed over the years, often under certain conditions on the kernels and functional characteristics around $x^*$ (see Sec.~\ref{sec:related_work} for a detailed discussion). Surprisingly, despite the long history, an algorithm with guaranteed order-optimal regret performance  remains open as discussed in~\citet{Vakili2022Open}.   \\

GP-based BO under noisy query was studied much more recently, following the pioneering work by ~\citet{Srinivas2010GPUCB} where they proposed the celebrated GP-UCB algorithm. Extensive studies since then have fully characterized the achievable learning performance, both in terms of information-theoretic lower bounds~\cite{Scarlett2017LowerBoundGP} and the design of algorithms such as SupKernel-UCB~\cite{Valko2013SupKernelUCB}, GP-ThreDS~\cite{Salgia2021GPThreDS}, BPE~\cite{Li2021BatchedPureExp}, and RIPS~\cite{Camilleri2021RIPS} that achieve the optimal performance.  \\

Under both the noise-free and noisy settings, a key practical concern for GP-based algorithms is their computational cost. The major computational bottleneck of prevailing GP-based algorithms is the maximization of an \emph{acquisition} function for choosing the query point at each time instant. The acquisition functions are often non-convex and computationally expensive to maximize. To achieve low regret order, such an optimization often needs to be carried out with increasing accuracy as time goes, resulting in a high overall computational requirement.

\subsection{Main Results}
\label{sub:main_results}

We explore a new design methodology for GP-based BO: an open-loop exploration of the domain using query points sampled at random from an arbitrary probability distribution supported over the domain. We show that this random exploration approach, while simplistic in nature, leads to 
order-optimal regret guarantees under both noise-free and noisy feedback models, thus closing the long standing regret gap in the noise-free setting. Moreover, the non-adaptive nature of random sampling bypasses the expensive step of optimizing a non-convex acquisition function, offering a computationally efficient solution without sacrificing learning efficiency. \\

Random exploration, while not new to many problems (see Sec.~\ref{sec:related_work}), has not been considered or analyzed for GP-based BO. It stands in sharp contrast to the prevailing exploratory query strategy in GP-based BO: the maximum posterior variance (MPV) sampling. Under MPV, the learning algorithm at each time queries the point with the highest posterior variance conditioned on past observations, i.e., a greedy approach to maximal uncertainty reduction. Surprisingly, we show that the simple, non-adaptive scheme of random exploration achieves the same order of predictive performance as MPV sampling, which is known to be order-optimal. In particular, we show that the worst-case posterior variance corresponding to $n$ randomly drawn points is bounded with high probability by $\tilde{\cO}(\gamma_{n}/n)$ and $\tilde{\cO}(n^{1- \beta})$ under noisy and noise-free feedback models, where $\gamma_n$ is the maximal information gain from $n$ query points and $\beta > 1$ is the order of the polynomial eigendecay of the kernel (see Sec.~\ref{sec:problem_statement} for their definitions). \\ 
 
A simpler solution is often more demanding when it comes to establishing optimality in performance. The drastically different nature of random exploration from MPV demands different analytical techniques in characterizing its predictive performance. The tightest bound on the worst-case predictive error of MPV sampling, derived in~\citet{Wenzel2021}, was obtained using the results on scattered data interpolation (i.e., approximating an unknown function using a given set of points) of functions in Sobolev spaces that provide bounds on the worst-case estimation error of the best interpolant based on the fill distance of the given set of points~\cite{Wendland2004, Narcowich2006SobolevEE, Brenner2008FEM, Arcangeli2012, Wenzel2021}. Since RKHSs of Mat\'ern kernels are norm-equivalent to Sobolev spaces, these results also immediately translate to estimation errors for function interpolation in RKHSs. The analytical techniques used in these studies require various technical assumptions on the regularity of the function domain and its boundary. These technical assumptions on the function domain present major challenges in incorporating MPV sampling with effective optimization techniques such as domain shrinking/elimination, hindering its potential applicability in designing algorithms with optimal regret. In contrast, in analyzing random exploration, we establish the concentration of the spectrum of the sample covariance operator to that of the true covariance operator that holds \emph{universally} for all compact domains. The crux of our analysis builds upon a careful treatment of the infinite-dimensional operators to separately ensure the concentration of the initial spectrum (consisting of the larger eigenvalues) and the tail spectrum, which allows us to obtain optimal convergence rate. The simplicity of random exploration in its implementation and the generality in its guaranteed predictive performance as established in this work make this exploration strategy an attractive alternative to MPV. We believe that the tools and techniques established here are of independent interest for extending the methodology of random exploration to other problem fields. \\

Built upon the above key results on random exploration, we develop and analyze a new algorithm for GP-based BO. Referred to as Random Exploration with Domain Shrinking (REDS),  this algorithm integrates the exploration strategy of random sampling with the optimization technique of domain shrinking~\cite{Li2021BatchedPureExp, Salgia2021GPThreDS}. Under the noise-free feedback model, we show that REDS incurs a cumulative regret of $\tilde{\cO}(\max\{T^{(3-\beta)/2}, 1\})$, which closes the gap to the known lower bound established in~\citet{Tuo2020Kriging} and hence resolves the longstanding open problem. The generality of random exploration, both in terms of the design methodology and performance guarantee is the reason behind the optimal regret performance of REDS. In particular, the order-optimal predictive performance of random exploration that holds universally over all compact domain enables a seamless integration of this exploration strategy with domain shrinking. Similarly, in the noisy setting, we show that REDS offers a cumulative regret of $\tilde{\cO}(\sqrt{T\gamma_T})$, which is order-optimal up to logarithmic factors. \\

The computational advantage of REDS is evident due to the simplicity of random exploration. We further demonstrate this with empirical studies where we compare REDS with BPE~\cite{Li2021BatchedPureExp} and GP-ThreDS~\cite{Salgia2021GPThreDS}, all offering optimal regret performance.  GP-ThreDS was shown to be computationally more efficient than prevailing algorithms such as GP-UCB. We show that REDS offers a significant speed-up in running time over both algorithms without compromising the regret performance. As shown in Table~\ref{table:time_taken}, REDS offers a $\sim15\times$ and $\sim100\times$  speed-up in runtime over GP-ThreDS and BPE, respectively.

\subsection{Related Work}
\label{sec:related_work}

For GP-based BO with noise-free feedback, a number of algorithms such as GP-EI~\cite{Mockus1975}, EGO~\cite{Jones1998EGO}, knowledge-gradient policy~\cite{Frazier2008knowledge}, and GP-PI~\cite{Kushner1964, Torn1989GPPI, Jones2001Taxonomy} have been proposed, which have since become classical. We refer the reader to the excellent tutorial by~\citet{Brochu2010Tutorial} for a more detailed description of the classical approaches. Despite their good empirical performance and popularity, theoretical guarantee on the convergence of these algorithms has only been established relatively recently.~\citet{Vazquez2010Convergence} showed that EI converges almost surely for any function drawn from a GP prior of finite smoothness.~\citet{Grunewalder2010regret} established the convergence rate of a computationally infeasible version of EI. Later,~\citet{Bull2011GPEI} established convergence rates for the computationally feasible version, showing that GP-EI achieves the optimal \emph{simple} regret for Mat\'ern kernels with smoothness $\nu < 1$, which does not translate to optimal cumulative regret performance. More recently,~\citet{DeFreitas2012ExponentialRegret} proposed the Branch and Bound algorithm that achieves a constant cumulative regret in Bayesian setting under additional assumptions on the differentiability of the kernel and the behaviour around the unique global maximum, which in practice are difficult to verify. In contrast, REDS requires no such additional assumptions and is analyzed in the frequentist setting.~\citet{Lyu2020efficient} showed that for kernels with a polynomial eigendecay with parameter $\beta$ (See Definition~\ref{definition:polynomial_eigendecay}), the GP-UCB algorithm achieves a regret of $\cO(T^{\frac{1 + \beta}{2\beta}})$, which is sub-optimal, as shown in~\citet{Vakili2022Open}. \\

The idea of using random sampling has been explored in related fields. The reconstruction of square integrable functions using random samples is a well-studied problem~\cite{Bohn2017Error, Bohn2017Thesis, Bohn2018SparseGrid, Smale2004Shannon, Cohen2013Stability, Chkifa2015Discrete, Cohen2017Optimal}. In particular, a series of studies considers efficient reconstruction of functions in RKHS using random samples drawn from the domain~\cite{Kammerer2021Worst, Krieg2021FunctionApproxPart1, Krieg2021FunctionApproxPart2, Moeller2021SamplingRKHS}. Despite certain similarities in the problem setup, an important point of distinction is that these studies focus on bounding the $L_2$ error of the reconstruction. In this work, we focus on bounding the sup-norm (or equivalently, $L_{\infty}$ norm) of the estimation error, which is larger than the $L_2$ norm and more challenging than bounding the $L_2$ norm. Since the analysis of algorithms requires a bound on the sup-norm of the estimation error, existing results are not applicable here.

\section{Problem Statement}
\label{sec:problem_statement}

\subsection{RKHS and Mercer's Theorem}
\label{sub:RKHS}

Let $\cX$ be a compact subset of $\R^d$ and $\varrho$ a finite Borel measure supported on $\cX$. A measure $\varrho$ is said to be \emph{supported} on $\cX$ if $\varrho(\cY) > 0$ for all open sets $\cY \subset \cX$. For $\cX \subset \R^d$, this is equivalent to $\varrho$ being \emph{absolutely continuous} w.r.t. the Lebesgue measure. Let $L_2(\varrho, \cX)$ denote the Hilbert space of (real) functions defined over $\cX$ that are square-integrable w.r.t. $\varrho$\footnote{To be rigorous, each $f \in L_2(\varrho, \cX)$ represents the class of functions that are equivalent $\varrho$-everywhere.}. \\

Consider a positive definite kernel $k: \cX \times \cX \to \R$. A Hilbert space $\cH_k$ of functions on $\cX$ equipped with an inner product $\ip{\cdot}{\cdot}_{\cH_k}$ is called a Reproducing Kernel Hilbert Space (RKHS) with reproducing kernel $k$ if the following conditions are satisfied: (i) $\forall \ x \in \cX$, $k(\cdot, x) \in \cH_k$; (ii) $\forall \ x \in \cX$, $\forall \ f \in \cH_k$, $f(x) = \ip{f}{k(\cdot, x)}_{\cH_k}$. For simplicity, we use $\psi_x$ to denote $k(\cdot, x)$. The inner product induces the RKHS norm, $\|f\|_{\cH_k}^2 = \ip{f}{f}_{\cH_k}$. WLOG, we assume that $k(x,x) = \|\psi_x\|_{\cH_k}^2 \leq 1$. For brevity, we drop the subscript of $\cH_k$ from the inner product for the rest of the paper. \\

Mercer's Theorem provides an alternative representation for RKHSs through the eigenvalues and eigenfunctions of a kernel integral operator defined over $L_2(\varrho, \cX)$ using the kernel $k$.
\begin{theorem}{~\citep[Theorem 4.49]{Steinwart2008MercerTheorem}}
    Let $\cX$ be a compact metric space, $k: \cX \times \cX \to \R$ be a continuous kernel and $\varrho$ be a finite Borel measure supported on $\cX$. Then, there exists an orthonormal system of functions $\{\varphi_j\}_{j \in \N}$ in $L_2(\varrho, \cX)$ and a sequence of non-negative values $\{\lambda_j\}_{j \in \N}$ satisfying $\lambda_1 \geq \lambda_2 \geq \dots \geq 0$, such that 
    $\displaystyle k(x, x') = \sum_{j \in \N} \lambda_j \varphi_j(x) \varphi_j(x')$
    holds for all $x, x' \in \cX$ and the convergence is absolute and uniform over $x, x' \in \cX$. 
    Moreover, $\{(\lambda_j, \varphi_j)\}_{j \in \N}$ corresponds to the eigensystem of the kernel integral operator $T_k : L_2(\varrho) \to L_2(\varrho)$ given by $T_kf = \int_{\cX} k(\cdot, x) f(x) d\varrho(x)$ for all $f \in L_2(\varrho)$.
\end{theorem}
Consequently, the Mercer representation~\citep[Thm. 4.51]{Steinwart2008MercerTheorem} of the RKHS of $k$ is given as
\begin{align*}
    \cH_k = \left\{ f := \sum_{j \in \N} \alpha_j {\lambda_j}^{\frac{1}{2}} \varphi_j : \|f\|_{\cH_k}^2 = \sum_{j \in \N} \alpha_j^2 < \infty \right\}.
\end{align*}
This also implies that $\{\upsilon_j\}_{j \in \N}$ with $\upsilon_j = \sqrt{\lambda_j} \varphi_j$ is an orthonormal basis for $\cH_k$. The following definition characterizes a class of kernels based on their eigendecay profile corresponding to their Mercer representation.
\begin{definition}     \label{definition:polynomial_eigendecay}
    Let $\{\lambda_j\}_{j \in \N}$ denote the eigenvalues of a kernel $k$ arranged in the descending order. The kernel $k$ is said to satisfy the polynomial eigendecay condition with a parameter $\beta > 1$ if, for some universal constant $C > 0$, we have $\lambda_j \leq C j^{-\beta}$ for all $j \in \N$.
\end{definition}
The above class of kernels encompasses a large number of kernels including the widely used Mat\'ern family. We make the following assumption on the kernel $k$ which is commonly adopted in the literature~\cite{Vakili2020InfoGain, Chatterji2019Online, Riutort2023Practical}.
\begin{assumption} \label{ass:eigenfunction_bound}
    The eigenfunctions $\{\varphi_j\}_{j \in \N}$ corresponding to $k$ are continuous and hence bounded on $\cX$, i.e., there exists $F > 0$ such that $\sup_{x \in \cX} |\varphi_j(x)| \leq F$ for all $j \in \N$.
\end{assumption}

\subsection{Problem Formulation}
\label{sub:problem_formulation}

We consider the problem of optimizing a fixed and unknown function $f: \cX \to \R$, where $\cX \subset \R^d$ is a compact domain and $f \in \cH_k$ with $\|f\|_{\cH_k} \leq B$. A sequential optimization algorithm chooses a point $x_t \in \cX$ at each time $t$ and observes $y_t = f(x_t) + \varepsilon_t$. In the noise-free setting, $\varepsilon_t \equiv 0$ for all~$t$. For the noisy setting, we assume that $\{\varepsilon_t\}_{t = 1}^T$ are independent, zero-mean, $R$-sub Gaussian random variables for some fixed constant $R \geq 0$, i.e., $\E[\exp(\zeta \varepsilon_t)] \leq \exp(\zeta^2R^2/2)$, for all $\zeta \in \R$ and $t \leq T$. The performance of the sequential algorithm is measured using the notion of cumulative regret, as defined in Eqn.~\eqref{eqn:regret_definition}.

\subsection{Preliminaries on Gaussian Processes}

Under the GP model, the unknown function $f$ is treated hypothetically as a realization of $\text{GP}(0, k)$, a Gaussian Process over $\cX$ with zero mean and $k(\cdot, \cdot)$ as the covariance kernel. The noise terms $\varepsilon$ are also viewed as zero mean Gaussian variables with variance $\tau$. The conjugate property of GPs with Gaussian noise allows for a closed form expression of the posterior distribution. Specifically, let $\cZ_t = \{(x_i, y_i)\}_{i = 1}^t$ denote a collection of points and their corresponding observations obtained according to the model described in Sec.~\ref{sub:problem_formulation}. Then, conditioned on $\cZ_t$, the posterior distribution of $f$ is also a GP with the following mean and covariance functions:
\begin{align}
    \mu_{t, \tau}(x) & = k_{X_t, x}^{\top}(K_{X_t,X_t} + \tau I_{t})^{-1} Y_t, \label{eqn:posterior_mean} \\
    k_{t, \tau}(x, \bar{x}) & = k(x,\bar{x}) - k_{X_t, x}^{\top}(K_{X_t,X_t} + \tau I_{t})^{-1}k_{X_t, \bar{x}}, \label{eqn:posterior_variance} 
\end{align}
where $k_{X_t, x} = [k(x_1, x), \dots k(x_t, x)]^{\top}$, $Y_t = [y_1, \dots, y_t]^{\top}$, $K_{X_t, X_t} = [k(x_i, x_j)]_{i,j=1}^t$ and $I_t$ is the $t \times t$ identity matrix. The posterior variance at a point $x$ is given as $\sigma_{t, \tau}^2(x) = k_{t, \tau}(x,x)$. The expression for posterior mean and variance in the noise-free setting is simply obtained by setting $\tau = 0$ in the above relations. \\

The posterior mean and variance computed using the GP model above are powerful tools to predict the values of the unknown function $f$ and to quantify the uncertainty in the prediction. In particular, the prediction error at a point $x \in \cX$, $|f(x)-\mu_{t, \tau}(x)|$, can be upper bounded by $\alpha \sigma_{t, \tau}(x)$, for a certain scaling factor $\alpha > 0$ that depends on the feedback model~\cite{Vakili2021OptimalSimpleRegret}. \\

Lastly, we define the information gain of a set of points $X_n = \{x_1, x_2, \dots, x_n\}$ as
\begin{align}
    \tilde{\gamma}_{X_n, \tau} := \frac{1}{2}\log\left( \det\left(I_t + \tau^{-1} K_{X_n, X_n} \right) \right).
    \label{eqn:info_gain}
\end{align}
Similarly, we define the maximal information gain as $\gamma_{n, \tau} := \sup_{X_n \subset \cX^n} \tilde{\gamma}_{X_n, \tau}$. Maximal information gain is an important term that corresponds to the effective dimension of the kernel and helps characterize the regret of the algorithms. It depends only on the kernel and~$\tau$.

\section{The Predictive Performance of Random Exploration}
\label{sec:random_exploration}

The following theorem characterizes the predictive variance, and consequently the predictive error, of a set of randomly sampled points from the domain.

\begin{theorem}
    Let $\cX$ be a compact subset of $\R^d$, $\varrho$ be a finite Borel measure supported on $\cX$,  and $k: \cX \times \cX \to \R$ be a continuous kernel satisfying the polynomial eigendecay condition with parameter $\beta > 1$ (Defn.~\ref{definition:polynomial_eigendecay}). Let $X_n = \{x_1, x_2, \dots, x_n\}$ denote a collection of $n$ i.i.d. points drawn from $\cX$ according to $\varrho$. Let $\sigma_{n, 0}^2$ and $\sigma_{n, \tau}^2$ denote, respectively, the posterior variance conditioned on $X_n$ in the noise-free setting and the noisy setting with a noise variance of $\tau > 0$. Then, for a given $\delta \in (0,1)$, there exists a constant $\overline{N}(\delta, k, \varrho, \tau) > 0$, such that, with probability at least $1 - \delta$, for all $n > \overline{N}(\delta, k, \varrho, \tau)$,
    \begin{align*}
        \sup_{x \in \cX} \sigma_{n, \tau}^2(x) & = \cO\left(\frac{\tau\gamma_{n,\tau}}{n}\right) = \tilde{\cO}((n/\tau)^{\frac{1}{\beta} - 1}), \\
        \sup_{x \in \cX} \sigma_{n, 0}^2(x) & = \tilde{\cO}(n^{1- \beta}).
    \end{align*}
    \label{thm:concentration_random_sampling}
\end{theorem}
The above obtained bounds on the worst-case posterior variance under the random exploration scheme are order-optimal (up to polylogarithmic factors), matching the existing lower bounds~\cite{Scarlett2017LowerBoundGP, Tuo2020Kriging}. The above theorem also improves upon the best known results for noisy scattered data approximation. In particular, for the class of Mat\'ern kernels with smoothness $\nu$ (i.e., $\beta = (2\nu + d)/d$), Theorem~\ref{thm:concentration_random_sampling} implies a worst-case predictive error of $\tilde{\cO}(n^{-\frac{\nu}{2\nu + d}})$, improving upon the bound of $\tilde{\cO}(n^{-\frac{\nu}{2\nu + 2d}})$ established by~\citet[Corollary 3]{Wynne2021ConvergenceMisspecified}. \\

The constant $\overline{N}(\delta, k, \varrho, \tau)$ is related to the kernel $k$ and measure $\varrho$ through two fundamental functions, $N(R)$ and $T(R)$, which are given as follows for any $R \in \N$:
\begin{align*}
    N(R) & := \sup_{x \in \cX} \sum_{j = 1}^{R} \varphi_j^2(x), \\
    T(R) & := \sup_{x \in \cX} \sum_{j = R + 1}^{\infty} \lambda_j \varphi_j^2(x) = \sup_{x \in \cX} \sum_{j = R + 1}^{\infty} \upsilon_j^2(x).
\end{align*}
They are referred to as the spectral functions of the kernel (see~\citet{Grochenig2020Sampling} and references therein) because of their dependence on the eigensystem corresponding to the kernel $k$ induced by the measure $\varrho$. Both $N(R)$ and $T(R)$ are fundamental quantities that appear in the analysis of reconstruction and estimation of functions in general $L_2$ spaces. The function $N(R)$ corresponds to the inverse of the infimum of the Christoffel function~\cite{Dunkl2014OrthogonalPolynomials} in the special case of reconstruction using orthogonal polynomials. Under Assumption~\ref{ass:eigenfunction_bound} and the condition of polynomial eigendecay (Def.~\ref{definition:polynomial_eigendecay}), $\overline{N}(\delta, k, \varrho, \tau)$ can be shown to be bounded as $\cO(\max\{F^4, (F^2/\tau)^{\frac{1}{\beta - 1}}\}\log(F/\delta))$. The dependence of $\overline{N}(\delta, k, \varrho, \tau)$ on $\delta$ is mild, as evident from the previous expression. Lastly, $\overline{N}(\delta, k, \varrho, \tau)$ is inversely proportional to $\tau$. Note that Theorem~\ref{thm:concentration_random_sampling} ensures that a smaller value of $\tau$ results in a tighter bound on the posterior variance, which in turn requires a larger number of samples. We refer the interested reader to the Appendix~\ref{appendix:proof_of_theorem_concentration} for a more detailed discussion of $\overline{N}(\delta, k, \varrho, \tau)$ and its dependence on $N(R)$ and $T(R)$. For brevity, we drop the arguments and use the notation $\overline{N}$ in the rest of the paper. \\

We provide a sketch of the proof of Theorem~\ref{thm:concentration_random_sampling} below and refer the reader to Appendix~\ref{appendix:proof_of_theorem_concentration} for a detailed proof.
\begin{proof}
    The main idea of the proof is to relate the worst-case posterior variance conditioned on $X_n$ to $\tilde{\gamma}_{X_n, \tau}$. This relation is established in two parts. In the first part, we establish that as the number of samples grow, the spectrum of random operator $\hat{\mathbf{Z}}$ concentrates to that of $\mathbf{Z}$, where $\hat{\mathbf{Z}}, \mathbf{Z}: \cH_k \to \cH_k$ are defined as follows:
    \begin{align*}
        \hat{\mathbf{Z}}g  := \left[\sum_{i = 1}^n \ip{g}{\psi_{x_i}} \psi_{x_i}\right] + \tau g; \quad \mathbf{Z}  := \E_{X_n}[\hat{\mathbf{Z}}],
    \end{align*}
    where $\{x_1, x_2, \dots, x_n\}$ denotes the random ensemble of points drawn according to the measure $\varrho$. The concentration in spectral norm allows us to approximate the expression of $\sigma_{n, \tau}^2(x) = \tau \langle{\psi_x},{\hat{\mathbf{Z}}^{-1}\psi_x}\rangle$ as $\sigma_{n, \tau}^2(x) \approx \tau \langle{\psi_x},{{\mathbf{Z}}^{-1}\psi_x}\rangle$, i.e., by replacing the sample covariance operator, $\hat{\mathbf{Z}}$, with the true covariance operator, $\mathbf{Z}$. Here, $A^{-1}$ denotes the inverse of an operator $A$, i.e., $A \circ A^{-1} = A^{-1} \circ A = \Id$ and $\Id$ denotes the identity operator. Thus, this step allows us to obtain a deterministic bound on posterior variance, which is easier to understand and analyze. We establish the required relation using the following two lemmas:
    \begin{lemma} \label{lemma:l2_norm}
        For all $n \geq \overline{N}$, the following relation holds with probability $1 - \delta/2$:
        \begin{align*}
            \|\mathbf{Z}^{-\frac{1}{2}}\hat{\mathbf{Z}}\mathbf{Z}^{-\frac{1}{2}} - \Id\|_2 \leq 1/9.
        \end{align*}
    \end{lemma}
    \begin{lemma} \label{lemma:l2_to_operator_norm}
        If the relation $\|\mathbf{Z}^{-\frac{1}{2}}\hat{\mathbf{Z}}\mathbf{Z}^{-\frac{1}{2}} - \Id\|_2 \leq b$ is true for some $b \in (0,1/3)$, then following is true $\forall \ x \in \cX$:
        \begin{align*}
            \langle{\psi_x},{\hat{\mathbf{Z}}^{-1}\psi_x}\rangle \leq \frac{\sqrt{1 - b}}{\sqrt{1 - b} - \sqrt{2b}} \cdot \langle{\psi_x},{{\mathbf{Z}}^{-1}\psi_x}\rangle.
        \end{align*}
    \end{lemma}
    Lemma~\ref{lemma:l2_norm} forms the cornerstone of the proof of the theorem. The result is established 
    by bounding the expression $|\ip{g}{(\mathbf{Z}^{-1/2}\hat{\mathbf{Z}}\mathbf{Z}^{-1/2} - \Id)g}|$  for an arbitrary $g$ with $\|g\|_{\cH_k} = 1$. We bound the above expression by decomposing it into a sum of three terms. Each of the three terms is then carefully bounded using a combination of Matrix-Chernoff inequality~\citep[Theorem 1.1]{Tropp2012TailBoundsMatrices}, a result for spectral norm concentration based on non-commutative Khinchtine inequality~\cite{Buchholz2001NonCommutativeKhintchine, Buchholz2005OptimalCI, Moeller2021SamplingRKHS} and Bernstein inequality. Lemma~\ref{lemma:l2_to_operator_norm} is established using a combination the structure of covariance matrices, the Cauchy-Schwarz inequality and the relation between the operator norm and $2$-norm. We would like to emphasize that both the above lemmas are true in general for \emph{all} eigendecay profiles and even without Assumption~\ref{ass:eigenfunction_bound} being true. \\

    In the second part, we show that, with high probability, the information gain of the (random) set $X_n$ is lower bounded by $n \cdot \sup_{x \in \cX} \langle{\psi_x},{{\mathbf{Z}}^{-1}\psi_x}\rangle$, upto a multiplicative constant.  The above idea is formalized in the following lemma.
    \begin{lemma} \label{lemma:information_gain}
        For all $n \geq \overline{N}$, the following relation holds with probability $1 - \delta/2$:
        \begin{align*}
            \tilde{\gamma}_{X_n, \tau} \geq \frac{13}{54F^2} \cdot n \cdot \sup_{x \in \cX} \langle{\psi_x},{{\mathbf{Z}}^{-1}\psi_x}\rangle.
        \end{align*}
    \end{lemma}
    Thus $\langle{\psi_x},{{\mathbf{Z}}^{-1}\psi_x}\rangle$ serves as the bridge for connecting the posterior variance to maximal information gain. \\

    The result for the noisy case follows immediately from the above lemmas by noting that $\gamma_{X_n, \tau} \leq \gamma_{n, \tau}$. For the noise-free setting, the results do not carry forward immediately as the above analysis does not hold for $\tau = 0$. To circumvent this issue, we use the fact that $\sigma_{n, \tau}^2(x)$ is an increasing function of $\tau$. Thus, we obtain a bound on $\sigma_{n, 0}^2(x)$ by using the bound on $\sigma_{n, \tau^*}^2(x)$, where $\tau^*$ is a carefully chosen value that not only allows us to use the analysis from the noisy case but also ensures that $\sigma_{n, \tau^*}^2$ is a close representation of $\sigma_{n, 0}^2$ to guarantee tightest possible bounds. 
    \end{proof}
    
\begin{remark}
    We would like to emphasize that the above result holds for samples generated under \emph{every} finite Borel measure $\varrho$ supported on $\cX$. However, the quality of the estimate changes with the choice of the measure through the leading constant in the bound in Theorem~\ref{thm:concentration_random_sampling}. 
\end{remark}

\section{The REDS algorithm}
\label{sec:REDS}

In this section, we present the proposed algorithm and analyze its regret performance. 

\subsection{REDS with Noise-Free Feedback}

REDS integrates random exploration with domain shrinking. It proceeds in epochs, maintaining an active region $\cX_r$ of the domain during each epoch $r \geq 1$. The sequence of active regions $\{\cX_r\}_{r}$ shrinks across epochs, i.e., $\cX_r \subseteq \cX_{r - 1} \subseteq \dots \cX_1 = \cX$, while ensuring $x^* \in \cX_r$ for all $r$ with high probability. During the $r^{\text{th}}$ epoch, REDS samples $N_r$ points, uniformly at random from the set $\cX_r$\footnote{If $\cX_r$ consists of multiple disjoint regions, then we carry out this step for each region separately.}, where $N_r = N_1 \cdot 2^{r-1}$ and the initial batch size $N_1$ is an input to the algorithm. \\

Using the observations from these points, REDS computes the posterior mean and variance function over $\cX_r$, denoted by $\mu_{r}$ and $\sigma^2_{r}$ respectively, using the Equations~\eqref{eqn:posterior_mean} and~\eqref{eqn:posterior_variance} with $\tau = 0$. The posterior mean and variance are then used to obtain $\cX_{r+1}$, an improved localization of $x^*$, as follows:
\begin{align*}
    \cX_{r+1} = \left\{ x \in \cX_r \ \bigg| \ \UCB_r(x) \geq \sup_{x' \in \cX_{r}} \LCB_r(x')  \right\}.
\end{align*}
Here, $\UCB(x) = \mu_r(x) + B\sigma_r(x)$ and $\LCB(x) = \mu_r(x) - B\sigma_r(x)$ correspond to upper and lower bounds on the estimate of $f$. A pseudocode for the algorithm is provided in Algorithm~\ref{alg:REDS}.

\begin{algorithm}[ht]
    \caption{Random Exploration with Domain Shrinking}
    \label{alg:REDS}
    \begin{algorithmic}[1]
            \STATE \textbf{Input}: $N_1$, the initial batch size.
            \STATE Set $\cX_1 \leftarrow \cX$, $t_{\text{curr}} \leftarrow 0$, $r \leftarrow 1$ 
            \FOR{$t = t_{\text{curr}} + 1, t_{\text{curr}} + 2, \dots, t_{\text{curr}} + N_r$}
            \STATE Sample a point $x_t$ uniformly at random from $\cX_r$ and observe $y_t$
            \IF{ $t > T$} 
            \STATE Terminate
            \ENDIF
            \ENDFOR
            \STATE Construct $\mu_{r}$ and $\sigma_{r}$ based on observations $\{(x_t, y_t: t \in \{t_{\text{curr}} +1, t_{\text{curr}} + 2, \dots, N_r\}\}$ using Eqn~\eqref{eqn:posterior_mean} and~\eqref{eqn:posterior_variance} with $\tau = 0$.
            \STATE Set $\cX_{r+1} = \{ x \in \cX_r \ | \ \UCB_{r}(x) \geq \sup_{x' \in \cX_r} \LCB_{r}(x') \}$
            \STATE $t_{\text{curr}} \leftarrow t_{\text{curr}} + N_r$, $N_{r+1} \leftarrow 2N_r$
            \STATE $r \leftarrow r+1$
    \end{algorithmic}
\end{algorithm}

\subsection{REDS under noisy feedback}

The REDS algorithm can be extended to operate under noisy feedback with the following two minor modifications to Algorithm~\ref{alg:REDS}. First, the posterior mean and variance $(\mu_{r, \tau},~\sigma_{r, \tau}^2)$ in each epoch should be computed using a noise variance $\tau > 0$ (Line $9$ of Algorithm~\ref{alg:REDS}). Second, the upper and lower confidence bounds, i.e., UCB and LCB (Line $10$ of Algorithm~\ref{alg:REDS}), should be updated to the following:
\begin{align}
    \UCB_{r, \tau, \delta}(x) & := \mu_{r, \tau}(x) + \alpha_{\tau, \delta} \sigma_{r, \tau}(x) + c_{T, \tau, \delta} \label{eqn:noisy_UCB} \\
    \LCB_{r, \tau, \delta}(x) & := \mu_{r, \tau}(x) - \alpha_{\tau, \delta} \sigma_{r, \tau}(x) - c_{T, \tau, \delta},  \label{eqn:noisy_LCB}
\end{align}
where $\alpha_{\tau, \delta} = B + R\sqrt{(2/\tau)\log(|\cD_T|/\delta)}$, $c_{T, \tau, \delta} = \frac{2B}{T} + R \sqrt{\frac{2}{T\tau}\log\left(\frac{4T}{\delta} \right)}$ and $\cD_T$ is defined in Assumption~\ref{ass:discretization}.

\subsection{Performance Analysis}

For the analysis of the REDS algorithm, we need to make the following two additional assumptions.

\begin{assumption} \label{ass:discretization}
    For all $n \in \N$, there exists a discretization $\cD_n$ of $\cX$ such that for all $f \in \cH_k$, $|f(x) - f([x]_{\cD_n})| \leq \|f\|_{\cH_k}/n$ and $|\cD_n| = \text{poly}(n)$\footnote{The notation $f(x) = \mathrm{poly}(x)$ is equivalent to $f(x) = \cO(x^k)$ for some $k \in \mathbb{N}$.}, where $[x]_{\cD_n} = \argmin_{y \in \cD_n} \|x - y\|_2$, is the point in $\cD_n$ that is closest to $x$.
\end{assumption}

\begin{assumption}
    Let $\cL_{\eta} = \{x \in \cX | f(x) \geq \eta \}$ denote the level set of $f$ for $\eta \in [-B, B]$. We assume that for all $\eta \in [-B,B]$, $\cL_{\eta}$ is a disjoint union of at most $M_f < \infty$ components, each of which is closed and connected. Moreover, for each such component, there exists a bi-Lipschitzian map\footnote{We refer the reader to the supplementary material for additional details about the terms used in this assumption.} between each such component and $\cX$ with normalized Lipschitz constant pair $L_f, L_f' < \infty$.
    \label{assumption:f_level_set_regularity}
\end{assumption}

Assumption~\ref{ass:discretization} is only required for the noisy case and is a standard assumption adopted in the literature. The existence of such a discretization has been justified and adopted in previous studies~\cite{Srinivas2010GPUCB, Chowdhury2017IGPUCB, Vakili2021OptimalSimpleRegret, SalgiaPersonalization2022} and is a mild assumption on the kernel. Specifically, the popular class of kernels like Squared Exponential and Mat\'ern kernels are known to be Lipschitz continuous, in which case a $\varepsilon$-cover of the domain with $\varepsilon = \cO(1/n)$ is sufficient to show the existence of such a discretization. Assumption~\ref{assumption:f_level_set_regularity} is an assumption on the regularity of the level sets of the function $f$. The existence of a bi-Lipschitzian map between two sets implies topological similarity between the two sets. Intuitively, this assumption ensures that the shape of the level-sets is not ``too arbitrary''. Note that such an assumption on the level sets of $f$ is relatively mild as the RKHS endows smoothness properties to the function $f$ which translate to a degree of topological regularity of level sets~\cite{Giovanni2011SardTheorem, Lee2010TopologicalManifolds}. \\

The following theorem characterizes the regret performance of REDS under noise-free feedback. 

\begin{theorem} \label{thm:noiseless_regret}
    Assume that the kernel $k$ satisfies the polynomial eigendecay condition with parameter $\beta > 1$ and function $f$ satisfies Assumption~\ref{assumption:f_level_set_regularity}. For a given $\delta \in (0,1)$, if REDS algorithm is run with $N_1 \geq C_{L_f, L_f'} \overline{N}(\delta/\log_2(T))$ and noise-free feedback, then the regret incurred by REDS satisfies,  
    \begin{align*}
        R(T) = \tilde{\cO}(\max\{T^{\frac{3-\beta}{2}}, 1\}).
    \end{align*}
    with probability at least $1 - \delta$. Here, $C_{L_f, L_f'}$ is a constant that depends only on $L_f$ and $L_f'$.
\end{theorem}
The following is an immediate corollary of the above theorem for the case of Mat\'ern kernels. 
\begin{corollary} \label{corollary:noise_free}
    Let $k$ be the Mat\'ern kernel with smoothness $\nu > 0$. For a given $\delta \in (0,1)$, if REDS algorithm is run with $N_1 \geq C_{L_f, L_f'}\overline{N}(\delta/\log_2(T))$ under noise-free feedback on a function $f \in \cH_k$ satisfying Assumption~\ref{assumption:f_level_set_regularity}, then the regret incurred by REDS satisfies,  
    \begin{align*}
        R(T) = \begin{cases} \tilde{\cO}(T^{1 - \nu/d}) & \text{ if } \nu < d, \\ \cO((\log T)^{5/2}) & \text{ if } \nu = d, \\ \cO((\log T)^{3/2}) & \text{ if } \nu > d.\end{cases}.
    \end{align*}
    with probability at least $1 - \delta$. Here, $C_{L_f, L_f'}$ is a constant that depends only on $L_f$ and $L_f'$.
\end{corollary}
This matches the result conjectured in~\citet{Vakili2022Open} upto logarithmic factors, \emph{resolving the open problem}. \\

The following theorem characterizes the regret performance of REDS in the noisy feedback setting.

\begin{theorem} \label{thm:noisy_regret}
    Consider the noisy observation model described in Sec.~\ref{sub:problem_formulation} and assume that Assumptions~\ref{ass:discretization} and~\ref{assumption:f_level_set_regularity} hold. For a given $\delta \in (0,1)$, if REDS algorithm is run with $N_1 \geq C_{L_f, L_f'} \overline{N}(\delta/(2\log_2 T))$ and UCB and LCB functions as defined in Eqns.~\eqref{eqn:noisy_UCB} and~\eqref{eqn:noisy_LCB} with parameter $\delta' = \delta/(2 \log_2 T)$, then the regret incurred by REDS satisfies, 
    \begin{align*}
        R(T) = \tilde{\cO}(\sqrt{T\gamma_T} \log(T/\delta)).
    \end{align*}
    with probability at least $1 - \delta$.
\end{theorem}

As shown by the above theorem, REDS achieves order-optimal regret (upto logarithmic factors) even under the noisy feedback model. \\

The proofs of both Theorems~\ref{thm:noiseless_regret} and~\ref{thm:noisy_regret} follow a similar blueprint. A key aspect of both the proofs is to ensure that as Theorem~\ref{thm:concentration_random_sampling} is invoked across the sets $\{\cX_r\}_{r \in \N}$, the leading constant in Theorem~\ref{thm:concentration_random_sampling}, which has an implicit dependence on the domain through the constant $F$, remains bounded and is independent of $T$. The following lemma shows that for all functions $f$ satisfying Assumption~\ref{assumption:f_level_set_regularity}, the leading constant only depends on the function and the initial domain.
\begin{lemma}
    Let $f \in \cH_k$ be such that Assumption~\ref{assumption:f_level_set_regularity} holds. Let $\cX'$ denote a path connected component of any level set of $f$ and $X' \subset \cX'$ be a set of $n$ points drawn uniformly at random from $\cX'$. Then for $n \geq C_{{L}, {L}_f'} \overline{N}(\delta)$, the following relations holds with probability $1 - \delta$: 
    \begin{align*}
        \sup_{x \in \cX'} \sigma_{X', \tau}^2(x) & \leq C_{{L}, {L}_f'}' \cdot F^2 \tau \cdot \frac{\gamma_{n, \tau}}{n} \\
        \sup_{x \in \cX'} \sigma_{X', 0}^2(x) & \leq C_{{L}, {L}_f'}' \cdot F^2 \cdot n^{1 -\beta}
    \end{align*}
    where $F$ and $\overline{N}(\delta)$ represent, respectively, the constants in Assumption~\ref{ass:eigenfunction_bound} and Theorem~\ref{thm:concentration_random_sampling}  corresponding to the uniform measure on $\cX$, and $C_{{L}, {L}_f'}, C_{{L}_f, {L}_f'}'$ are constants that depend only on ${L}_f, {L}_f'$. 
    \label{lemma:variance_across_domains}
\end{lemma}
At a high level, the above lemma ensures that under the regularity condition on the topology of level sets (Assumption~\ref{assumption:f_level_set_regularity}), Theorem~\ref{thm:concentration_random_sampling} can be applied across level sets of $f$ by just paying the penalty of a constant that depends only on $f$. The proof is based on the inclusion of RKHSs over subsets along with a change of measure argument. We refer the reader to Appendix~\ref{appendix:proof_of_regret_theorems} for a detailed proof of Lemma~\ref{lemma:variance_across_domains} and Theorems~\ref{thm:noiseless_regret} and~\ref{thm:noisy_regret}.

\section{Empirical Studies}
\label{sec:simulation}

\begin{figure*}[t]
\centering
\subfloat[Branin]{\label{fig:branin}\centering \includegraphics[scale = 0.25]{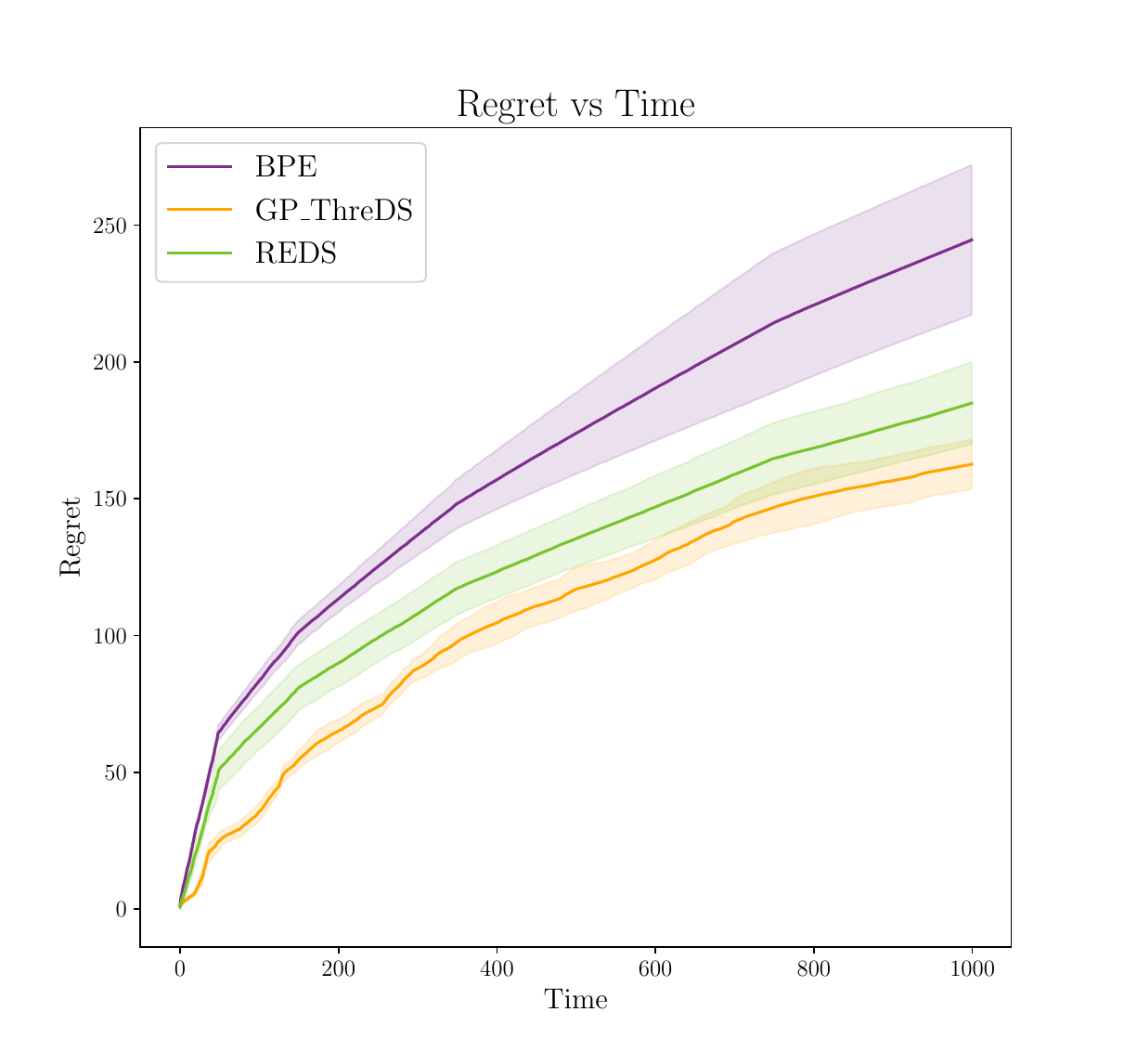}}
~
\subfloat[Hartmann-4D]{\label{fig:hartmann_4D}\centering \includegraphics[scale = 0.265]{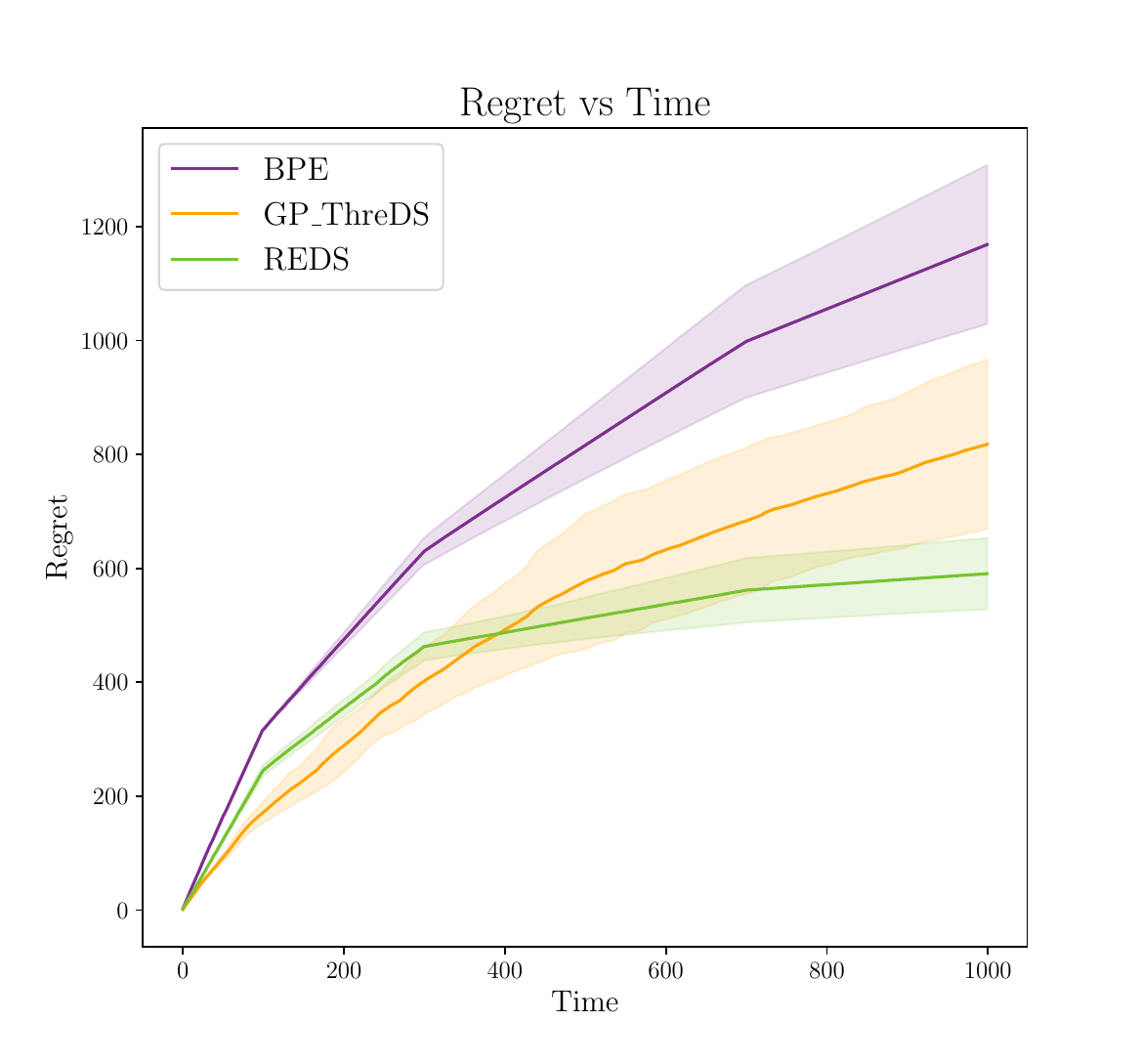}}
~
\subfloat[Hartmann-6D]{\label{fig:hartmann_6D}\centering \includegraphics[scale = 0.255]{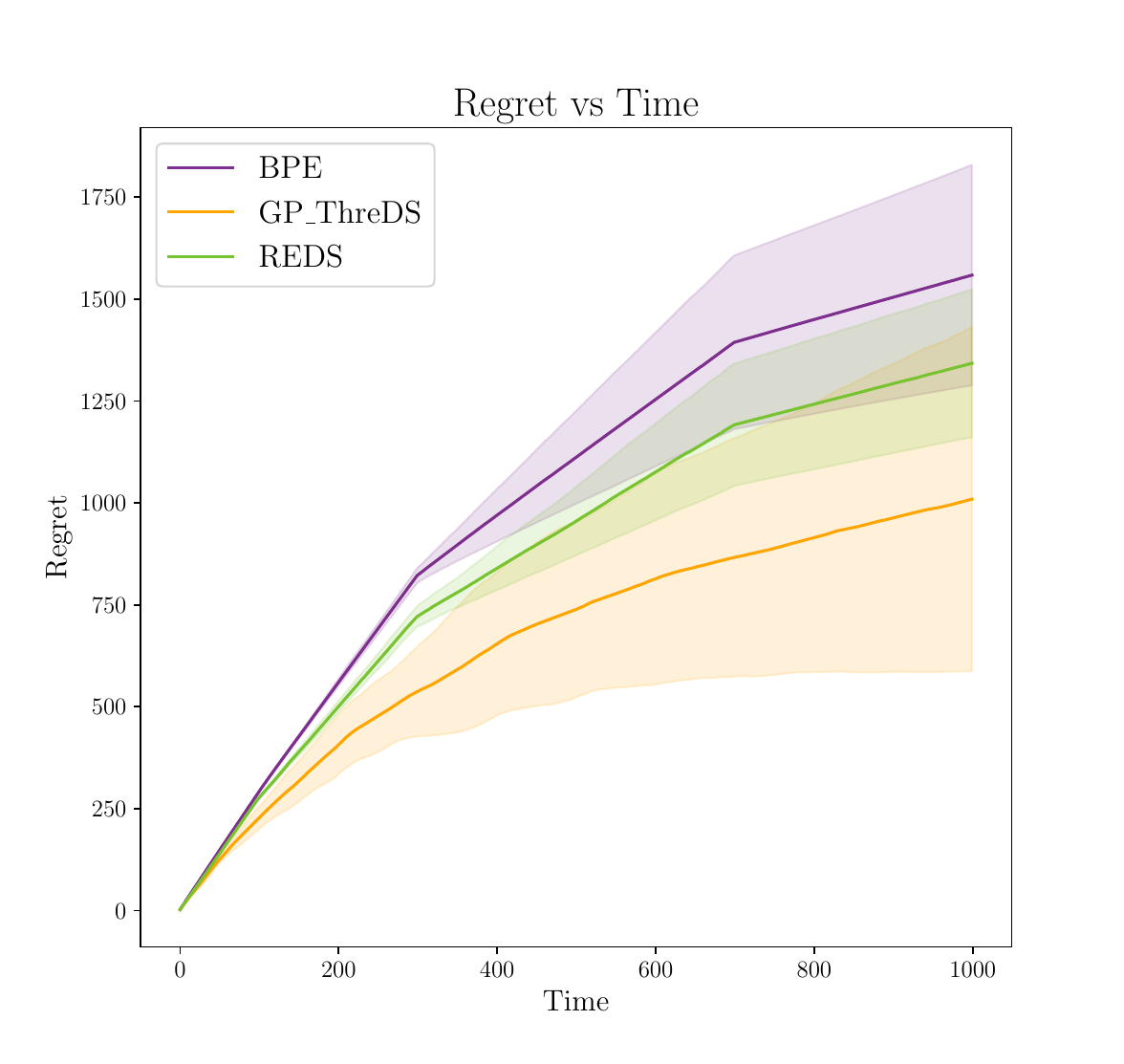}}
~
\caption{Cumulative regret averaged over $10$ Monte Carlo runs for all algorithms across different benchmark functions. The shaded region represents the error bars upto one standard deviation. As evident from the plots, the regret of REDS is comparable to that of BPE and GP-ThreDS.}
\label{fig:regret_plots}
\end{figure*}

We compare the computational efficiency of REDS against algorithms with order-optimal regret performance, namely BPE~\citep{Li2021BatchedPureExp} and GP-ThreDS~\citep{Salgia2021GPThreDS} through an empirical study. We compare the regret performance and the running time of the three algorithms for three commonly used benchmark functions in Bayesian Optimization, namely, Branin~\citep{Azimi2012Branin, Picheny2013Branin}, Hartmann-4D~\citep{Picheny2013Branin} and Hartmann-6D~\citep{Picheny2013Branin}. The analytical expressions for the three benchmark functions are given as follows:
\begin{itemize}
    \item Branin function, denoted by $B(x_1, x_2)$, is defined over $\cX = [0,1]^2$.
    \begin{align*}
        B(x_1, x_2) & = -\frac{1}{51.95}\left( \left( v - \frac{5.1u^2}{4 \pi^2} + \frac{5u}{\pi} - 6\right)^2 +  \left(10 - \frac{10}{8\pi} \right)\cos(u) - 44.81\right),
    \end{align*}
     where $u = 15x_1 - 5$ and $v = 15x_2$.
     \item Hartmann-$4$D function, denoted by $H_4(x_1, x_2, x_3, x_4)$, is defined over $\cX = [0,1]^4$.
     \begin{align*}
         H_4(x_1, x_2, x_3, x_4) = \sum_{i = 1}^4 w_i \exp\left( - \sum_{j = 1}^4 A_{ij}(x_j - C_{ij})^2\right).
     \end{align*}
     \item Hartmann-$6$D function, denoted by $H_6(x_1, x_2, x_3, x_4, x_5, x_6)$, is defined over $\cX = [0,1]^6$.
     \begin{align*}
         H_6(x_1, x_2, x_3, x_4, x_5, x_6) = \sum_{i = 1}^4 w_i \exp\left( - \sum_{j = 1}^6 A_{ij}(x_j - C_{ij})^2\right).
     \end{align*}
\end{itemize}
In the definitions above, $w_i$ denotes the $i^{\text{th}}$ element of the vector $w = \begin{pmatrix} 1.0 & 1.2 & 3.0 & 3.2 \end{pmatrix}^{\top}$ and
$A_{ij}$ and $C_{ij}$ refer to the $(i,j)^{\text{th}}$ element of the matrices $A$ and $C$, defined below:
\begin{align*}
    A = \begin{pmatrix} 10 & 3 & 17 & 3.5 & 1.7 & 8 \\ 0.05 & 10 & 17 & 0.1 & 8 & 14 \\3 & 3.5 & 1.7 & 10 & 17 & 8  \\ 17 & 8 & 0.05 & 10 & 0.1 & 14\end{pmatrix}; \quad C  = 10^{-4} \cdot \begin{pmatrix} 1312 & 1696 & 5569 & 124 & 8283 & 5886 \\ 2329 & 4135 & 8307 & 3736 & 1004 & 9991 \\ 2348 & 1451 & 3522 & 2883 & 3047 & 6650 \\ 4047 & 8828 & 8732 & 5743 & 1091 & 381 \end{pmatrix}
\end{align*}

For BPE and REDS, we consider a discretized version of the domain consisting of $2000$, $7000$ and $20000$ points chosen uniformly at random from the domain for the Branin, Hartmann-$4$D and Hartmann-$6$D functions respectively. We use the exponentially growing epoch schedule for both BPE and REDS as described in (Algorithm~\ref{alg:REDS}) for a fair comparison. We implement GP-ThreDS as described in~\citet{Salgia2021GPThreDS}. For each node in the tree, we consider a discretization, chosen uniformly at random, of size $100$, $200$ and $500$ for the Branin, Hartmann-$4$D and Hartmann-$6$D functions respectively. The values of $(a,b)$ (the lower and upper bound on $f(x^*)$) are set to $(0.5, 1.2)$, $(0, 3.8)$ and $(0, 3.5)$ for Branin, Hartmann-$4$D and Hartmann-$6$D respectively. We set $\tau = 0.2$ for all experiments. The value of $\alpha_{\tau}$ is set to $1$ across all experiments, except for BPE with Hartmann-$4$D and Hartmann-$6$D for which we set it to $0.75$. These values are obtained using a grid search over $[0.25, 2]$ in steps of $0.25$. The parameter $N_1$ in REDS and BPE was set to $50$ for Branin and $100$ for Hartmann-$4$D and Hartmann-$6$D functions. \\

\begin{table} 
    \centering
    \begin{tabular}{cccc}
    \toprule
         & BPE  & GP-ThreDS & REDS \\ \midrule
       Branin  & $29.84 \pm 6.13$ & $4.37 \pm 0.28$   & $\mathbf{0.32} \pm 0.08$ \\ \midrule
       Hartmann-4D & $38.45 \pm 3.93$ & $7.59 \pm 0.54$ & $\mathbf{0.47} \pm 0.11$\\ \midrule
       Hartmann-6D & $119.71 \pm 23.75$   & $19.33 \pm 0.54$  & $\mathbf{1.19} \pm 0.08$  \\ \bottomrule
    \end{tabular}
    \caption{Time taken (in seconds) by different algorithms across the different benchmark functions.}
    \label{table:time_taken}
\end{table}

For all the experiments, we used the Square exponential kernel. The length scale was set to $0.2$ for Branin and $1$ for Hartmann-$4$D and Hartmann-$6$D functions. We corrupted the observations with a zero mean Gaussian noise to the with a standard deviation of $0.2$. All the algorithms were run for $T = 1000$ time steps. We recorded the cumulative regret and time taken by different algorithms for $10$ Monte Carlo runs for each benchmark function. \\

The regret for the algorithms over different functions is plotted in Figure~\ref{fig:regret_plots}. The shaded region represents the error bars upto standard deviation on either side. The running times, with an error bar of one standard deviation, are tabulated in Table~\ref{table:time_taken}. As evident from the plots in Figure~\ref{fig:regret_plots}, the regret incurred by REDS is comparable to that of other algorithms for all benchmark functions. At the same time, REDS offers about a $15\times$ and $100\times$ speedup in terms of runtime over the GP-ThreDS and BPE (See Table~\ref{table:time_taken}), demonstrating the practical benefits of our proposed methodology of random sampling.

\section{Conclusion}
\label{sec:conclusion}

In this work, we studied the methodology of exploring the domain using random samples drawn from a distribution supported on a compact domain. We showed that this non-adaptive approach offers the optimal-order of worst case predictive error for RKHS function in both noisy and noise-free feedback settings. The proposed approach offers a simple alternative for designing Bayesian Optimization algorithms which typically involve choosing points through a computationally expensive step of optimizing a non-convex acquisition function. Based on this methodology, we developed a algorithm  that achieves order-optimal regret in both noisy and noise-free settings, \emph{resolving a COLT open problem}. We demonstrated the computational advantage of the proposed approach through an empirical study, where the proposed algorithm achieved upto a $100\times$ runtime speed up over state-of-the-art algorithms.

\bibliography{references}
\bibliographystyle{abbrvnat}

\newpage
\appendix
\section{Proof of Theorem~\ref{thm:concentration_random_sampling}}
\label{appendix:proof_of_theorem_concentration}

We begin with setting up some notation that will be used throughout the proof. Throughout the appendix, we will represent the elements of $\cH_k$ as infinite dimensional vectors and operators over these function spaces as infinite dimensional matrices. We adopt such a convention for ease for presentation while keeping in mind that despite the matrix representation, the actual operation is over elements of $\cH_k$. Recall that we defined the sample covariance operator $\hat{\bfZ}$ for a randomly chosen sample $X_n = \{x_1, x_2, \dots, x_n\}$ and its expected value $\bfZ = \E[\hat{\bfZ}]$ as follows for any $g \in \cH_k$:
\begin{align*}
    \hat{\bfZ}g &  := \left[\sum_{i = 1}^n \ip{g}{\psi_{x_i}} \psi_{x_i}\right] + \tau g \\
     \bfZ & := \E[\hat{\bfZ}].
\end{align*}
In the matrix-vector notation, the operators (equivalently, matrices) are given as:
\begin{align*}
    \hat{\bfZ} & := \left(\sum_{i = 1}^n \psi_{x_i} \psi_{x_i}^{\top}\right) + \tau \Id \\
    \bfZ & = \E[\hat{\bfZ}] = \E\left[\sum_{i = 1}^n \psi_{x_i} \psi_{x_i}^{\top}\right] + \tau \Id \\
    & = n \E[\psi_{x_1} \psi_{x_1}^{\top}] + \tau \Id = n \bfLambda + \tau \Id,
\end{align*}
where $\Id$ is the identity matrix (operator) and $\bfLambda = \text{diag}(\lambda_1, \lambda_2, \dots)$ is the diagonal matrices consisting of the eigenvalues of the kernel $k$ corresponding to the measure $\varrho$. If we define $\bfPsi_n := [\psi_{x_1}, \psi_{x_2}, \dots, \psi_{x_n}]$, then we can also write $\hat{\bfZ} = \bfPsi_n \bfPsi_n^{\top} + \tau \Id$. Consequently, the posterior variance at any point $x \in \cX$ is given as:
\begin{align*}
    \sigma_{n, \tau}^2(x) = \tau \psi_{x}^{\top} \hat{\bfZ}^{-1} \psi_x.
\end{align*}

For any $R \in \N$, we define the following two quantities that will be relevant during our analysis:
\begin{align}
    N(R) & := \sup_{x \in \cX} \sum_{j = 1}^{R} \varphi_j^2(x), \\
    T(R) & := \sup_{x \in \cX} \sum_{j = R + 1}^{\infty} \lambda_j \varphi_j^2(x) = \sup_{x \in \cX} \sum_{j = R + 1}^{\infty} \upsilon_j^2(x).
\end{align}
Recall that $\{\varphi_j\}_{j \in \N}$ are eigenfunctions of the kernel operator and form an orthonormal system in $L_2(\varrho, \cX)$ and $\{\upsilon\}_j$ are an orthonormal basis for $\cH_k$. The term $N(R)$ is often referred to as the spectral function (see~\cite{Grochenig2020Sampling} and references therein) and in case of orthogonal polynomials, it is the inverse of the infimum of the Christoffel function~\cite{Dunkl2014OrthogonalPolynomials}. Both $N(R)$ and $T(R)$ are fundamental quantities that appear in the analysis of reconstruction and estimation of functions. \\

Lastly, based on $N(R)$ and $T(R)$, for a given kernel $k$, measure $\varrho$ and $\delta \in (0,1)$, we define the following terms for any $n \in \N$ and $\tau > 0$:
\begin{align*}
    \cR_{k, \varrho}^{(1)}(n, \tau, \delta) & := \left\{ R \in \N : N(R) \leq \frac{n}{1944 \log(6n/\delta)}\right\} \\
    \cR_{k, \varrho}^{(2)}(n, \tau, \delta) & := \left\{ R \in \N : \max\{ 42T(R), n \lambda_{R+1} \} \log \left(\frac{12}{\delta} \right)  \leq \frac{\tau}{27}\right\}\\
    \cR_{k, \varrho}(n, \tau, \delta) & := \cR_{k, \varrho}^{(1)}(n, \tau, \delta) \cap \cR_{k, \varrho}^{(2)}(n, \tau, \delta) \\
    \overline{N}(k, \varrho,\delta, \tau) & := \max\left\{\min \left\{n : \cR_{k, \varrho}(n, \tau, \delta) \neq \emptyset \right\}, \lceil 729 \cdot F^4 \cdot \log(12/\delta) \rceil \right\}
\end{align*}
The dependence on $k$ and $\varrho$ is implicit through $\{\varphi_j\}_{j \in \N}$ and $\{\lambda_j\}_{j \in \N}$ used to define $N(R)$ and $T(R)$. For brevity of notation, going forward, we drop the explicit description of dependence on $k$ and $\varrho$. \\

We are now ready to prove the theorem. We first prove the statement of the theorem, assuming that the lemmas hold, followed by the proofs of the lemmas. \\ 

We begin with result for the noisy case, where $\tau > 0$ is fixed (independent of $n$). From Lemma~\ref{lemma:l2_norm}, we know that for $n \geq \overline{N}$, $\|\bfZ^{-1/2}\hat{\bfZ}\bfZ^{-1/2} - \Id\|_2 \leq 1/9$ holds with probability $1 - \delta$. Using this result along Lemma~\ref{lemma:l2_to_operator_norm}, we can conclude that $\psi_x^{\top} \hat{\bfZ}^{-1} \psi_x \leq 2 \psi_x^{\top} {\bfZ}^{-1} \psi_x$ holds for all $x$. Thus, we have,
\begin{align}
    \sigma_{n, \tau}^2(x) & = \tau \psi_{x}^{\top} \hat{\bfZ}^{-1} \psi_x \nonumber \\
     & \leq 2\tau \psi_{x}^{\top} {\bfZ}^{-1} \psi_x \nonumber \\
     & \leq \frac{108F^2}{13} \cdot \tau \cdot \frac{\tilde{\gamma}_{X_n, \tau}}{n} \nonumber \\
     & \leq \frac{108F^2}{13} \cdot \tau \cdot \frac{\gamma_{n, \tau}}{n}, \label{eqn:sigma_noisy_bound}
\end{align}
as required. The third line in the above expression follows from Lemma~\ref{lemma:information_gain}. We would like to emphasize that the polynomial eigendecay condition is not necessary to obtain the above relation. It is only necessary to bound the information gain in terms on $n$. Under the polynomial eigendecay condition with parameter $\beta > 1$, the above equation can also be written as
\begin{align*}
    \sigma_{n, \tau}^2(x) \leq C_0 \cdot \left(\frac{n}{\tau}\right)^{\frac{1}{\beta} - 1} 
    \log(n),
\end{align*}
where we used the bound on information gain from~\citet[Corollary 1]{Vakili2020InfoGain} and $C_0$ is an appropriately chosen constant independent of $n$ and $\tau$. \\

We now consider the noise-free case. Since information gain is only defined for $\tau > 0$, we cannot directly extend the analysis as used in the noisy case by substituting $\tau = 0$. To circumvent this issue, we carefully choose $\tau^* > 0$, such that $\sigma_{n, \tau^*}^2$ is a close representation of $\sigma_{n, 0}^2$. We choose $\tau^*$ to be dependent on $n$ such that $\tau^*$ goes to $0$ as $n$ becomes larger. This allows $\sigma_{n, \tau^*}^2$ to faithfully represent the value of $\sigma_{n, 0}^2$ over the range of $n$. Specifically, we choose $\tau^* = c' n^{1- \beta} (\log (n/\delta))^{\beta}$ for $c' \geq C(1944 F^2)^{\beta}$, where $C$ is the constant in Assumption~\ref{ass:eigenfunction_bound}. The condition on constant $c'$ ensures that $\overline{N}(k, \varrho, \delta, \tau^*)$ exists. Since all conditions of the analysis for $\tau > 0$ (noisy case) are satisfied, we can directly invoke the result for $\tau > 0$. Using the bound on $\sigma_{n, \tau}^2$ and the monotonicity of $\sigma_{n, \tau}^2$ as a function of $\tau$, we obtain,
\begin{align*}
    \sigma_{n, 0}^2(x) \leq \sigma_{n, \tau^*}^2(x) \leq C_1 \cdot n^{1 - \beta} (\log (n/\delta))^{\beta},
\end{align*}
where $C_1$ is a constant independent of $n$. \\

In the following subsections, we prove Lemmas~\ref{lemma:l2_norm},~\ref{lemma:l2_to_operator_norm} and~\ref{lemma:information_gain}.

\subsection{Proof of Lemma~\ref{lemma:l2_norm}}
\label{appendix:proof_of_lemma_l2_norm}

Since we are interested in bounding the $2$-norm of the operator $\bfZ^{-1/2}\hat{\bfZ}\bfZ^{-1/2} - \Id$, we will focus on finding an upper bound on $g^{\top}(\bfZ^{-1/2}\hat{\bfZ}\bfZ^{-1/2} - \Id)g$ that holds uniformly for all functions $g$ in the unit ball in RKHS, i.e., $\{g: \|g\|_{\cH_k} \leq 1 \}$. The high level idea is to 
separately consider the contribution of component of $g$ that belongs to the subspace spanned by eigenfunctions corresponding to the ``large'' eigenvalues, i.e., head of the spectrum and those corresponding to the ``small'' eigenvalues, i.e., tail of the spectrum. \\

Throughout the proof, we fix a $R \in \cR_{n, \tau}$. The existence of such an $R$ is guaranteed by the assumption $n > \overline{N}$. For the analysis, we define two projection operators, $\bfP $ and $\bfQ $. We define $\bfP $ as the projection operator onto the subspace spanned by $\{\upsilon_j\}_{j = 1}^{R}$, i.e., for any $g = \sum_{j \in \N} g_j \upsilon_j \in \cH_k$, $\bfP g = \sum_{j = 1}^R g_j \upsilon_j$. Note that $\bfP $ is an orthogonal projection operator. Similarly, we define $\bfQ  = \text{Id} - \bfP $. \\

We also introduce some additional notation for the ease of presentation. We define $\bfL$ to be the diagonal matrix (operator) whose $j^{\text{th}}$ entry is $\dfrac{\lambda_j}{n\lambda_j + \tau}$. Similarly, let $\omega_i = \bfLambda^{-1/2}\psi_{x_i}$ for $i = 1,2,\dots, n$. Using this notation, we can rewrite the matrix $\bfZ^{-1/2}\hat{\bfZ}\bfZ^{-1/2} - \Id$ as
\begin{align*}
    \bfZ^{-1/2}\hat{\bfZ}\bfZ^{-1/2} - \Id & = \bfZ^{-1/2} \left( \sum_{i = 1}^{n} \psi_{x_i}\psi_{x_i}^{\top} + \tau \Id \right) \bfZ^{-1/2} -\Id \\
    & =   \sum_{i = 1}^{n} (\bfZ^{-1/2}\psi_{x_i})(\bfZ^{-1/2}\psi_{x_i})^{\top} + \tau \bfZ^{-1} -\Id \\
    & =  \sum_{i = 1}^{n} (\bfL^{1/2}\omega_{i})(\bfL^{1/2}\omega_{i})^{\top} - n\bfL.
\end{align*}

For any $g \in \cH_k$, we have the following decomposition:
\begin{align}
    |g^{\top}(\bfZ^{-1/2}\hat{\bfZ}\bfZ^{-1/2} - \Id)g| & = |(\bfP g + \bfQ g)^{\top}(\bfZ^{-1/2}\hat{\bfZ}\bfZ^{-1/2} - \Id)(\bfP g + \bfQ g)| \nonumber \\
    & \leq |(\bfP g)^{\top} (\bfZ^{-1/2}\hat{\bfZ}\bfZ^{-1/2} - \Id)(\bfP g)| + |(\bfQ g)^{\top} (\bfZ^{-1/2}\hat{\bfZ}\bfZ^{-1/2} - \Id) (\bfQ g)| \nonumber \\
    & ~~~~~~ + |(\bfP g)^{\top} (\bfZ^{-1/2}\hat{\bfZ}\bfZ^{-1/2} - \Id) + (\bfQ g)^{\top} (\bfZ^{-1/2}\hat{\bfZ}\bfZ^{-1/2} - \Id) (\bfP g)| \nonumber \\
    & \leq \underbrace{|g^{\top} \bfP (\bfZ^{-1/2}\hat{\bfZ}\bfZ^{-1/2} - \Id)\bfP  g|}_{:=E_1} + \underbrace{|g^{\top} \bfQ (\bfZ^{-1/2}\hat{\bfZ}\bfZ^{-1/2} - \Id)\bfQ  g|}_{:=E_2} \nonumber \\
    & ~~~~~~ + 2\underbrace{|g^{\top} \bfP (\bfZ^{-1/2}\hat{\bfZ}\bfZ^{-1/2} - \Id)\bfQ  g|}_{:= E_3}. \label{eqn:split_into_P_and_Q}
\end{align}

We separately bound the terms $E_1, E_2$ and $E_3$, beginning we $E_1$. We have,
\begin{align}
    E_1 & = |g^{\top} \bfP (\bfZ^{-1/2}\hat{\bfZ}\bfZ^{-1/2} - \Id)\bfP  g| \nonumber \\
    & = \left| (\bfP g)^{\top} \bfP \left( \sum_{i = 1}^{n} (\bfL^{1/2}\omega_{i})(\bfL^{1/2}\omega_{i})^{\top} - nL\bfL\right)\bfP  (\bfP g) \right| \nonumber \\
    & = \left| (\bfP g)^{\top} \left( \sum_{i = 1}^{n} (\bfP \bfL^{1/2}\bfP \omega_{i})(\bfP \bfL^{1/2} \bfP \omega_{i})^{\top} - n\bfP \bfL \bfP  \right) (\bfP g) \right| \nonumber\\
    & = n \left| (\bfP g)^{\top} \bfP \bfL^{1/2}\bfP  \left( \frac{1}{n}\sum_{i = 1}^{n} (\bfP \omega_{i})(\bfP \omega_{i})^{\top} - \bfP   \right)\bfP \bfL^{1/2}\bfP  (\bfP g) \right| \nonumber\\
    & \leq n \left\| \left(\frac{1}{n}\sum_{i = 1}^{n} (\bfP \omega_{i})(\bfP \omega_{i})^{\top} - \bfP \right) \right\|_{2}  \cdot \|\bfP \bfL^{1/2}\bfP  (\bfP g) \|_{\cH_k}^2 \nonumber\\
    & \leq n \left\| \left(\frac{1}{n}\sum_{i = 1}^{n} (\bfP \omega_{i})(\bfP \omega_{i})^{\top} - \bfP \right) \right\|_{2}  \cdot (g^{\top} \bfP \bfL \bfP  g) \nonumber \\
    & \leq  \left\| \left(\frac{1}{n}\sum_{i = 1}^{n} (\bfP \omega_{i})(\bfP \omega_{i})^{\top} - \bfP \right) \right\|_{2}  \cdot (n\|\bfL\|_2) \cdot \|\bfP g\|_{\cH_k}^2. \label{eqn:E_1_expansion}
\end{align}
In the above equations, we used the fact that for any diagonal matrix $D$, $\bfP D = D\bfP  = \bfP D\bfP $ and that $\bfP ^2 = \bfP $. Firstly, note that $\|\bfL\|_2 = \max_{j \in \N} \lambda_j/(n \lambda_j + \tau) \leq 1/n$. Consequently, $n \|\bfL\|_2 \leq 1$. Secondly, to bound the first term on the RHS, we denote $\bfP \omega_i := A_i$ for all $i = 1,2,\dots, n$. We have, $\E[A_i A_i^{\top}] = \bfP  \E[\omega_i \omega_i^{\top}] \bfP  = \bfP  \bfLambda^{-1/2} \E[\psi_{x_i} \psi_{x_i}^T] \bfLambda^{-1/2}\bfP  = \bfP  \bfLambda^{-1/2} \bfLambda \bfLambda^{-1/2}\bfP  = \bfP $. Moreover, for all $A_i$'s, only the top $R \times R$ sub-matrix has non-zero entries, implying it is sufficient to bound the $2$-norm of that finite sub-matrix to bound the first term on the RHS. We use Matrix-Chernoff inequality~\citep[Theorem 1.1]{Tropp2012TailBoundsMatrices} to bound the $2$-norm of this finite dimensional submatrix. \\

For all $i = 1,2, \dots, n$, let $[A_i]_{R} \in \R^R$ denote the $R$-dimensional vector corresponding to the first $R$ coordinates of $A_i$. Thus, we are interested in applying the Matrix-Chernoff inequality to bound the following expression:
\begin{align*}
    E_{11} := \left\| \left(\frac{1}{n}\sum_{i = 1}^{n} [A_i]_{R} [A_i]_R^{\top}  - I_{R}\right) \right\|_{2},
\end{align*}
where $I_{R}$ denotes the $R$ dimensional identity matrix. Here, we used the fact that the relevant $R \times R$ sub-matrix of $\bfP $, or equivalently $\E[[A_1]_{R} [A_1]_R^{\top}]$,  corresponds to $I_{R}$. To invoke the Matrix-Chernoff inequality, we need bounds on the maximum and minimum eigenvalue of $\displaystyle \E\left[ \frac{1}{n}\sum_{i = 1}^{n} [A_i]_{R} [A_i]_R^{\top} \right]$ and a bound on $\|[A_i]_{R} [A_i]_R^{\top}/n\|_2$ that holds almost surely for all $i = 1,2, \dots, n$. Since  $\E[[A_1]_{R} [A_1]_R^{\top}] = I_R$, $\displaystyle \E\left[ \frac{1}{n}\sum_{i = 1}^{n} [A_i]_{R} [A_i]_R^{\top} \right] = I_R$ implying that both the maximum and minimum eigenvalues are $1$. For any $i = 1, 2, \dots, n$, we have,
\begin{align*}
    \frac{\|[A_i]_{R} [A_i]_R^{\top}\|_2}{n} & \leq \frac{1}{n} \tr([A_i]_{R} [A_i]_R^{\top}) 
     \leq \frac{1}{n} \tr( [A_i]_R^{\top}[A_i]_{R}) 
      \leq \frac{1}{n} \|\bfP \omega_i\|_{\cH_k}^2 
      \leq \frac{1}{n} \sum_{j = 1}^R \varphi_j^2(x_i) 
      \leq \frac{N(R)}{n}.
\end{align*}
On invoking the Matrix-Chernoff inequality with these results, we obtain that the following relation is true with probability $1 - \delta/6$:
\begin{align}
    E_{11} \leq \sqrt{\frac{3N(R) \log(3R/\delta)}{n}}. \label{eqn:E_11_bound}
\end{align}
On combining the above bound with Eqn.~\eqref{eqn:E_1_expansion} along with noting that $n\|L\|_2 \leq 1$, we can conclude that:
\begin{align}
    E_1 \leq \sqrt{\frac{3N(R) \log(3R/\delta)}{n}} \cdot \|\bfP g\|_{\cH_k}^2. \label{eqn:E_1_final}
\end{align}
We would like to mention that the above bound is only valid when the RHS in Eqn.~\eqref{eqn:E_11_bound} is less than $1$. However, this condition is satisfied by the choice of $n > \overline{N}$. \\

We now consider the second term, $E_2$. We have, 
\begin{align}
    E_2 & = |g^{\top} \bfQ (\bfZ^{-1/2}\hat{\bfZ}\bfZ^{-1/2} - \Id)\bfQ  g| \nonumber \\
    & = \left| (\bfQ g)^{\top} \left( \sum_{i = 1}^{n} (\bfQ \bfL^{1/2}\omega_{i})(\bfQ \bfL^{1/2}\omega_{i})^{\top} - n\bfQ \bfL\bfQ  \right) (\bfQ g) \right| \\
    & \leq n \underbrace{\left\| \left(\frac{1}{n}\sum_{i = 1}^{n} (\bfQ \bfL^{1/2}\omega_{i})(\bfQ \bfL^{1/2}\omega_{i})^{\top} - \bfQ \bfL\bfQ \right) \right\|_{2}}_{:=E_{21}}  \cdot \|\bfQ g\|_{\cH_k}^2. \label{eqn:E_2_expansion}
\end{align}
Note that the term $E_{21}$ has a similar structure as $E_{11}$ except for the fact that $E_{21}$ involves infinite-dimensional vectors as opposed to finite-dimensional vectors.  Thus, to bound $E_{21}$ we use a result from~\citet[Proposition 3.8]{Moeller2021SamplingRKHS} which is spectral concentration inequality for infinite-dimensional vectors derived using non-commutative Khinchtine inequality~\cite{Buchholz2001NonCommutativeKhintchine, Buchholz2005OptimalCI, Moeller2021SamplingRKHS}. From Proposition $3.8$ in~\citet{Moeller2021SamplingRKHS}, we can conclude that the following relation holds with probability at least $1 - \delta/6$:
\begin{align}
    \left\| \left(\frac{1}{n}\sum_{i = 1}^{n} (\bfQ \bfL^{1/2}\omega_{i})(\bfQ \bfL^{1/2}\omega_{i})^{\top} - \bfQ \bfL \bfQ \right) \right\|_{2} \leq \max \left\{ \frac{42}{n} \log\left(\frac{12}{\delta}\right)B_1 , B_2  \right\}, \label{eqn:infinite_spectral_concentration}
\end{align}
where $B_1 = \max_{i = 1,2,\dots, n} \|\bfQ \bfL^{1/2}\omega_{i}\|_{\cH_k}^2$ and $B_2 = \|\bfQ \bfL\bfQ \|_2$. We can further bound the terms $B_1$ and $B_2$ as follows.
\begin{align*}
    B_1 & = \max_{i = 1,2,\dots, n} \|\bfQ \bfL^{1/2}\omega_{i}\|_{\cH_k}^2 = \max_{i = 1,2,\dots, n} \sum_{j = {R+1}}^{\infty} \frac{\lambda_j}{n\lambda_j + \tau} \varphi_j^2(x_i) \leq \sup_{x \in \cX} \frac{1}{\tau} \sum_{j = {R+1}}^{\infty} \lambda_j \varphi_j^2(x) = \frac{T(R)}{\tau} \\
    B_2 & = \|\bfQ \bfL \bfQ \|_2 = \max_{j \in \N, j > R} \frac{\lambda_j}{n\lambda_j + \tau} \leq \frac{\lambda_{R+1}}{\tau}.
\end{align*}
On plugging this into Eqn.~\eqref{eqn:infinite_spectral_concentration}, we obtain the following bound on $E_{21}$.
\begin{align}
    E_{21} \leq \frac{1}{\tau} \left\{ \frac{42}{n} \log\left(\frac{12}{\delta}\right) T(R) , \lambda_{R+1}  \right\}. \label{eqn:E_21_final}
\end{align}
Combining Eqn.~\eqref{eqn:E_2_expansion} and~\eqref{eqn:E_21_final} yields us,
\begin{align}
    E_2 \leq \frac{1}{\tau} \left\{ 42 \log\left(\frac{12}{\delta}\right) T(R) , n\lambda_{R+1}  \right\} \|\bfQ  g\|_{\cH_k}^2. \label{eqn:E_2_final}
\end{align}

We now move onto the third term, $E_3$, which contains the cross terms. For brevity of notation, we define $\zeta_i := \bfP \bfL^{1/2}\omega_i$ and $\xi_{i} := \bfQ \bfL^{1/2}\omega_i$ for all $i = 1,2,\dots, n$. Note that $\zeta_i^{\top} \xi_j = 0$ for all $i,j = 1,2,\dots, n$. Since $\bfP $ and $\bfQ$ commute with $\bfL$, a diagonal matrix, it is straightforward to note that $\bfP \bfL \bfQ   = 0$. Using this relation along with the definition of $\{\zeta_i\}_{i = 1}^n$ and $\{\xi_i\}_{i = 1}^n$, we can rewrite $E_3$ as follows:
\begin{align}
    E_3 & = |g^{\top} \bfP (\bfZ^{-1/2}\hat{\bfZ}\bfZ^{-1/2} - \Id)\bfQ   g| \nonumber\\
    & = \left|  g^{\top} \bfP \left( \sum_{i = 1}^{n} (\bfL^{1/2}\omega_{i})(\bfL^{1/2}\omega_{i})^{\top} - n \bfL\right)\bfQ   g \right| \nonumber \\
    & = \left|   \sum_{i = 1}^{n} (g^{\top} \bfP \bfL^{1/2}\omega_{i})(g^{\top} \bfQ  \bfL^{1/2}\omega_{i})^{\top}   \right| \nonumber \\
    & = \left|   \sum_{i = 1}^{n} \underbrace{(g^{\top}\zeta_{i})(g^{\top} \xi_i)}_{:= W_i}  \right|. \label{eqn:E_3_expansion}
\end{align}

We use Bernstein inequality to bound the sum of the random variables $W_i$, for which we need the values of $\E[W_i]$, $\E[W_i^2]$ and an upper bound on $|W_i|$ that holds almost surely. We begin with $\E[W_i]$. We have,
\begin{align}
    \E[W_i] = \E[(g^{\top}\zeta_{i})(g^{\top} \xi_i)] = g^{\top} \E[\zeta_i \xi_i^{\top}] g = 0. \label{eqn:W_i_mean}
\end{align}
For an upper bound on $|W_i|$, note that for any $g$ with $\|g\|_{\cH_k} = 1$, $|W_i|$ is maximized for the choice of $g = \psi_{x_i}$. Thus, 
\begin{align}
    |W_i| & = \|g\|_{\cH_k}^2 \left( \frac{g^{\top}\zeta_{i}}{\|g\|_{\cH_k}} \right) \left( \frac{g^{\top} \xi_i}{\|g\|_{\cH_k}} \right) \nonumber\\
    & \leq \|g\|_{\cH_k}^2 (\psi_{x_i}^{\top}\zeta_{i})(\psi_{x_i}^{\top} \xi_i) \nonumber \\
    & \leq  \|g\|_{\cH_k}^2 \|\zeta_i\|_{\cH_k}^2  \|\xi_i\|_{\cH_k}^2 \nonumber \\
    & \leq \|g\|_{\cH_k}^2 \cdot \left( \sum_{j = 1}^{R} \frac{\lambda_j}{n\lambda_j + \tau} \varphi_j^2(x_i) \right) \cdot \left( \sum_{j = R+1}^{\infty} \frac{\lambda_j}{n\lambda_j + \tau} \varphi_j^2(x_i)  \right) \nonumber \\
    & \leq \|g\|_{\cH_k}^2 \cdot \left( \frac{1}{n} \sum_{j = 1}^{R}  \varphi_j^2(x_i) \right) \cdot \left( \frac{1}{\tau} \sum_{j = R+1}^{\infty} \lambda_j \varphi_j^2(x_i)  \right) \nonumber \\
    & \leq \|g\|_{\cH_k}^2 \cdot \frac{N(R)}{n} \cdot \frac{T(R)}{\tau}. \label{eqn:W_i_abs_bound}
\end{align}
From the above expressions, we can also conclude that $|g^{\top}\zeta_{i}| \leq \|g\|_{\cH_k}\cdot \dfrac{N(R)}{n}$ and $|g^{\top}\xi_{i}| \leq \|g\|_{\cH_k} \cdot  \dfrac{T(R)}{\tau}$. We use these relations to obtain a bound on $\E[W_i^2]$. We have,
\begin{align}
     \E[W_i^2] & = \E[(g^{\top}\zeta_{i})^2(g^{\top} \xi_i)^2] \nonumber \\
    & \leq \|g\|_{\cH_k}^2 \cdot \min\left\{ \E\left[ (g^{\top}\zeta_{i})^2 \right] \left( \frac{T(R)}{\tau} \right)^2, \E\left[  (g^{\top} \xi_i)^2  \right] \left( \frac{N(R)}{n} \right)^2  \right\} \nonumber \\
    & \leq \|g\|_{\cH_k}^2 \cdot \min\left\{  (g^{\top}\bfP L\bfP g) \cdot \left( \frac{T(R)}{\tau} \right)^2,   (g^{\top}\bfQ  L\bfQ  g) \cdot \left( \frac{N(R)}{n} \right)^2   \right\} \nonumber \\
    & \leq \|g\|_{\cH_k}^2 \cdot \min\left\{  \frac{\|\bfP g\|_{\cH_k}^2}{n} \cdot \left( \frac{T(R)}{\tau} \right)^2,   \frac{\lambda_{R+1} \|\bfQ  g\|_{\cH_k}^2}{\tau} \cdot \left( \frac{N(R)}{n} \right)^2   \right\}. \label{eqn:W_i_variance}
\end{align}
In the last step, we used the bounds on $\|\bfL\|_2$ and $\|\bfQ  \bfL\bfQ  \|_2$ derived in the earlier part of the proof. Lastly, since $\E[W_i] = 0$, $\var(W_i) = \E[W_i^2]$. On applying Bernstein inequality~\citep[Lemma 7.37]{Wasserman2008Bernstein} using the relations from Eqns.~\eqref{eqn:W_i_mean},~\eqref{eqn:W_i_abs_bound} and~\eqref{eqn:W_i_variance}, we can conclude that the following relation holds with probability $1 - \delta/6$:
\begin{align}
    E_3 & = \left|   \sum_{i = 1}^{n} (g^{\top}\zeta_{i})(g^{\top} \xi_i)   \right| \nonumber \\
    & \leq \|g\|_{\cH_k} \cdot \sqrt{2n \log \left(\frac{6}{\delta}\right) \min\left\{  \frac{\|\bfP g\|_{\cH_k}^2}{n} \cdot \left( \frac{T(R)}{\tau} \right)^2,   \frac{\lambda_{R+1} \|\bfQ  g\|_{\cH_k}^2}{\tau} \cdot \left( \frac{N(R)}{n} \right)^2   \right\}}  \nonumber \\
    & ~~~~~~~~~~~~~~~~~~~~~~~ + \|g\|_{\cH_k}^2  \cdot \frac{2N(R)}{3n} \cdot \frac{T(R)}{\tau}  \cdot \log \left(\frac{6}{\delta}\right). \label{eqn:E_3_final}
\end{align}

On plugging the results from Eqns.~\eqref{eqn:E_1_final},~\eqref{eqn:E_2_final} and~\eqref{eqn:E_3_final} into Eqn.~\eqref{eqn:split_into_P_and_Q}, we obtain
\begin{align*}
    \|\bfZ^{-1/2}\hat{\bfZ}\bfZ^{-1/2} - \Id\|_2 & = \sup_{g : \|g\|_{\cH_k} \leq 1} |g^{\top} (\bfZ^{-1/2}\hat{\bfZ}\bfZ^{-1/2} - \Id) g| \\
    & \leq \sup_{g : \|g\|_{\cH_k} \leq 1} \bigg[ \sqrt{\frac{3N(R) \log(6R/\delta)}{n}} \|\bfP g\|_{\cH_k}^2 + \frac{1}{\tau} \max \left\{ 42 \log\left(\frac{12}{\delta}\right) T(R) , n\lambda_{R+1}  \right\} \|\bfQ  g\|_{\cH_k}^2  \\
    & ~~~~~~~~~~~~ + 2\|g\|_{\cH_k} \sqrt{2n \log \left(\frac{6}{\delta}\right) \min\left\{  \frac{\|\bfP f\|_{\cH_k}^2}{n} \cdot \left( \frac{T(R)}{\tau} \right)^2,   \frac{\lambda_{R+1} \|\bfQ  f\|_{\cH_k}^2}{\tau} \cdot \left( \frac{N(R)}{n} \right)^2   \right\}} \\
    & ~~~~~~~~~~~~~~~~~~~~ + \|g\|_{\cH_k}^2  \cdot \frac{4N(R)}{3n} \cdot \frac{T(R)}{\tau}  \cdot \log \left(\frac{6}{\delta}\right) \bigg] \\
    & \leq \bigg[ \sqrt{\frac{3N(R) \log(6R/\delta)}{n}} + \frac{1}{\tau} \max \left\{ 42 \log\left(\frac{12}{\delta}\right) T(R) , n\lambda_{R+1}  \right\}   \\
    & ~~~~ + 2 \sqrt{2n \log \left(\frac{6}{\delta}\right) \min\left\{  \frac{1}{n} \cdot \left( \frac{T(R)}{\tau} \right)^2,   \frac{\lambda_{R+1}}{\tau} \cdot \left( \frac{N(R)}{n} \right)^2   \right\}} + \frac{4N(R)T(R)}{3n\tau}\log \left(\frac{6}{\delta}\right) \bigg]
\end{align*}

On plugging in any value of $R \in \cR(n, \tau, \delta)$ and using the definition of $\cR_{n, \tau, \delta}$ along with the relation $n \geq \overline{N}$, we can conclude that $\|\bfZ^{-1/2} \hat{\bfZ} \bfZ^{-1/2} - \Id \|_2 \leq 1/9$ with probability at least $1 - \delta/2$. The overall probability on the bound is obtained using a union bound for the relations on $E_1$, $E_2$ and $E_3$.

\subsection{Proof of Lemma~\ref{lemma:l2_to_operator_norm}}
\label{appendix:proof_of_lemma_l2_to_operator_norm}

We begin the proof by showing that we can relate the $\psi_{x}^{\top}\hat{\bfZ}^{-1}\psi_x$ to $\psi_{x}^{\top}{\bfZ}^{-1}\psi_x$ through the operator norm of $\bfM := \hat{\bfZ}^{-1/2}(\bfZ - \hat{\bfZ})\bfZ^{-1/2} $. Specifically, we show if that operator norm of $\bfM$ is small, then $\psi_{x}^{\top}\hat{\bfZ}^{-1}\psi_x$ and $\psi_{x}^{\top}{\bfZ}^{-1}\psi_x$ are within a constant factor of each other. Lastly, we use the condition on $\|\bfZ^{-\frac{1}{2}}\hat{\bfZ}\bfZ^{-\frac{1}{2}} - \Id\|_2$ to bound the $\|\bfM\|_{\text{op}}$, the operator norm of $\bfM$, to obtain the required result. \\

We begin with considering the following expression.
\begin{align}
    \left|\psi_{x}^{\top}(\hat{\bfZ}^{-1} - \bfZ^{-1})\psi_x\right| & = \left| \psi_x^{\top}\hat{\bfZ}^{-1}(\bfZ - \hat{\bfZ})\bfZ^{-1}\psi_x \right|  \nonumber \\
    & = \left| \psi_x^{\top} \hat{\bfZ}^{-1/2} \cdot \hat{\bfZ}^{-1/2}(\bfZ - \hat{\bfZ})\bfZ^{-1/2} \cdot \bfZ^{-1/2}\psi_x \right| \nonumber \\
    & \leq \| \hat{\bfZ}^{-1/2} \psi_x \|_{\cH_k} \| {\bfZ}^{-1/2} \psi_x \|_{\cH_k} \| \hat{\bfZ}^{-1/2}(\bfZ - \hat{\bfZ})\bfZ^{-1/2}  \|_{\text{op}} \nonumber\\
    & \leq \sqrt{(\psi_{x}^{\top}\hat{\bfZ}^{-1}\psi_x)} \cdot \sqrt{(\psi_{x}^{\top} \bfZ^{-1}\psi_x)} \cdot  \|  \bfM  \|_{\text{op}}. \label{eqn:Z_hat_Z_relate_op_norm}
\end{align}
Consider the scenario where the relation $\|\bfM\|_{\text{op}} \leq c$ is satisfied for some $c \in (0,1)$. We claim that under this scenario, we have, $\psi_{x}^{\top}\hat{\bfZ}^{-1}\psi_x \leq (1 - c)^{-1} \cdot \psi_{x}^{\top}{\bfZ}^{-1}\psi_x$. To show this claim, we consider Eqn.~\eqref{eqn:Z_hat_Z_relate_op_norm}. If 
$\psi_{x}^{\top}{\bfZ}^{-1}\psi_x \geq \psi_{x}^{\top}\hat{\bfZ}^{-1}\psi_x$, the claim follows immediately. For the other case, we have,
\begin{align*}
     \psi_x \hat{\bfZ}^{-1} \psi_x  -  \psi_x {\bfZ}^{-1} \psi_x   & \leq  \sqrt{(\psi_{x}^{\top}\hat{\bfZ}^{-1}\psi_x)} \cdot \sqrt{(\psi_{x}^{\top} \bfZ^{-1}\psi_x)} \cdot c \\
     & \leq   \sqrt{(\psi_{x}^{\top}\hat{\bfZ}^{-1}\psi_x)} \cdot \sqrt{(\psi_{x}^{\top}\hat{\bfZ}^{-1}\psi_x)} \cdot c \\
     & \leq   c \cdot (\psi_{x}^{\top}\hat{\bfZ}^{-1}\psi_x) \\
    \implies \psi_x \hat{\bfZ}^{-1} \psi_x & \leq \left(\psi_x {\bfZ}^{-1} \psi_x\right) \cdot \frac{1}{1 - c},
\end{align*}
as claimed. Thus, it suffices to show that $\|\bfM\|_{\text{op}}$ is small. \\

To that effect, note that we can write the operator $\bfM$ as $\bfM = \hat{\bfZ}^{-1/2}\bfZ^{1/2} - \hat{\bfZ}^{1/2}\bfZ^{-1/2} = \bfC^{-1} - \bfC^{\top}$ where, $\bfC := \bfZ^{-1/2}\hat{\bfZ}^{1/2}$. Consequently, using the definition of operator norm yields us,
\begin{align}
    \|\bfM\|_{\text{op}}^2  = \|\bfM^{\top}\bfM\|_2 & = \| ((\bfC^{\top})^{-1} - \bfC)(\bfC^{-1} - \bfC^{\top}) \|_2 \nonumber \\
    & = \| (\bfC\bfC^{\top})^{-1} - \text{Id} + \bfC\bfC^{\top} - \text{Id} \|_2 \nonumber \\
    & \leq \| (\bfC\bfC^{\top})^{-1} - \text{Id} \|_2 + \| \bfC\bfC^{\top} - \text{Id} \|_2. \label{eqn:M_operator_norm}
\end{align}
From the definition of $\bfC$, we have $\| \bfC\bfC^{\top} - \text{Id} \|_2 = \|\bfZ^{-\frac{1}{2}}\hat{\bfZ}\bfZ^{-\frac{1}{2}} - \Id\|_2 \leq b$, from the given statement in the Lemma. Note that if $ \| \bfC\bfC^{\top} - \text{Id} \|_2 \leq b$ for some $b \in (0,1/3)$, then all eigenvalues of $\bfC\bfC^{\top}$ lie in the interval $[1 - b, 1 + b]$. This implies that all the eigenvalues of $(\bfC\bfC^{\top})^{-1}$ lie in the interval $[(1 + b)^{-1}, (1 - b)^{-1}]$. Hence, $\| (\bfC\bfC^{\top})^{-1} - \text{Id} \|_2 \leq b/(1-b)$. On combining this with Eqn.~\eqref{eqn:M_operator_norm}, we can conclude that if $\| \bfC\bfC^{\top} - \text{Id} \|_2 \leq b$, then $\|\bfM\|_{\text{op}} \leq \sqrt{2b/(1-b)} < 1$. On combining this with the previous claim that relates $\psi_{x}^{\top}\hat{\bfZ}^{-1}\psi_x$ to $\psi_{x}^{\top}{\bfZ}^{-1}\psi_x$ through $\|\bfM\|_{\text{op}}$, we arrive at the result.

\subsection{Proof of Lemma~\ref{lemma:information_gain}}
\label{appendix:proof_of_lemma_information_gain}

Similar to the analysis in Appendix~\ref{appendix:proof_of_lemma_l2_norm}, we fix an $R \in \cR(n, \tau, \delta)$ and define projection matrices $\bfP $ and $\bfQ  $ using the value of $R$ as defined in Appendix~\ref{appendix:proof_of_lemma_l2_norm}. We define the projection of the kernel operator $k(\cdot, \cdot)$ on the subspaces spanned by $\bfP $ and $\bfQ  $ as follows:
\begin{align*}
    k^{(\bfP )}(x,y) = \sum_{j = 1}^R \lambda_{j} \varphi_j(x) \varphi_j(y); \quad k^{(\bfQ)}(x,y) = k(x, y) - k^{(\bfP )}(x,y).
\end{align*}
Recall that $\tilde{\gamma}_{X_n, \tau}$ denotes the information gain corresponding to the randomly drawn set of points $X_n = \{x_1, x_2, \dots, x_n\}$. Similar to $K_{X_n, X_n}$, we also define $K^{(\bfP )}_{X_n, X_n}$ and $K^{(\bfQ)}_{X_n, X_n}$ as $K^{(\bfP )}_{X_n, X_n} = [k^{(\bfP )}(x_i, x_j)]_{i, j = 1}^n$ and $K^{(\bfQ)}_{X_n, X_n} = [k^{(\bfQ)}(x_i, x_j)]_{i, j = 1}^n$. It is straightforward to note that $K_{X_n, X_n} = K^{(\bfP )}_{X_n, X_n} + K^{(\bfQ)}_{X_n, X_n}$.  \\

We first derive some auxiliary results on the spectrum of $K^{(\bfP )}_{X_n, X_n}$ and $K^{(\bfQ)}_{X_n, X_n}$ which will be useful in the analysis later. Recall that we defined $\bfPsi_n := [\psi_{x_1}, \psi_{x_2}, \dots, \psi_{x_n}]$. We can also rewrite $K_{X_n, X_n}$, $K^{(\bfP )}_{X_n, X_n}$ and $K^{(\bfQ)}_{X_n, X_n}$ in terms of $\bfPsi_n$ as: $K_{X_n, X_n} = \bfPsi_n^{\top} \bfPsi_n$, $K^{(\bfP )}_{X_n, X_n} = \bfPsi_n^{\top} P\bfPsi_n$ and $K^{(\bfQ)}_{X_n, X_n} = \bfPsi_n^{\top} \bfQ   \bfPsi_n$. Using this relation, note that the singular values of $K^{(\bfP )}_{X_n, X_n} = (P\bfPsi_n)^{\top}(P\bfPsi_n)$ and $K^{(\bfQ)}_{X_n, X_n} = \bfPsi_n^{\top} \bfQ   \bfPsi_n$ are the same as that of $(P\bfPsi_n)(P\bfPsi_n)^{\top} = P\bfPsi_n \bfPsi_n^{\top}P$ and $(\bfQ  \bfPsi_n)(\bfQ  \bfPsi_n)^{\top} = \bfQ  \bfPsi_n \bfPsi_n^{\top}\bfQ  $ respectively. \\

For the spectrum of $K^{(\bfP )}_{X_n, X_n}$, note that 
\begin{align*}
    K^{(\bfP )}_{X_n, X_n} & = (\bfP  \bfPsi_n)^{\top}(\bfP  \bfPsi_n) = ((n\bfLambda)^{-1/2}\bfP  \bfPsi_n)^{\top} (n\bfLambda) ((n\bfLambda)^{-1/2}\bfP  \bfPsi_n) \\
    & = (\bfP  (n\bfLambda)^{-1/2}\bfPsi_n)^{\top} \bfP  (n\bfLambda) \bfP   (\bfP  (n\bfLambda)^{-1/2}\bfPsi_n).
\end{align*}
If $\tilde{\lambda}_1 \geq \tilde{\lambda}_2 \geq \dots \geq \tilde{\lambda}_{R}$ denote the eigenvalues of $K^{(\bfP )}_{X_n, X_n}$, then using Ostrowski's Theorem~\cite{Ostrowski1959Theorem}, we can conclude that $\tilde{\lambda}_j = \theta_j n\lambda_j$ for all $j = 1,2,\dots, R$, where $\{n\lambda_j\}_{j = 1}^R$ correspond to the eigenvalues of $n\bfP  \bfLambda \bfP  $ and $\theta_j$ lie between the smallest and largest eigenvalues of the matrix $n^{-1}(\bfP  \bfLambda^{-1/2}\bfPsi_n)^{\top} (\bfP  \bfLambda^{-1/2}\bfPsi_n)$. Note that the singular values (in this case, also eigenvalues) of $n^{-1}(\bfP  \bfLambda^{-1/2}\bfPsi_n)^{\top} (\bfP  \bfLambda^{-1/2}\bfPsi_n)$ are the same as that of $n^{-1}(\bfP  \bfLambda^{-1/2}\bfPsi_n)(\bfP  \bfLambda^{-1/2}\bfPsi_n)^{\top} = n^{-1}\sum_{i = 1}^n (\bfP  \omega_i) (\bfP  \omega_i)^{\top}$, where $\omega_i = \bfLambda^{-1/2}\psi_{x_i}$, as defined in Appendix~\ref{appendix:proof_of_lemma_l2_norm}. Using Eqn.~\eqref{eqn:E_11_bound} and that $R \in \cR(n, \tau, \delta)$ and $n \geq \overline{N}$, we can conclude that the following relation is true with probability $1 - \delta/6$:
\begin{align*}
    \left\| \left(\frac{1}{n}\sum_{i = 1}^{n} (\bfP  \omega_{i})(\bfP  \omega_{i})^{\top} - \bfP  \right) \right\|_{2} \leq \frac{1}{27}.
\end{align*}
Thus, we can conclude that eigenvalues of $n^{-1}(\bfP  \bfLambda^{-1/2}\bfPsi_n)^{\top} (\bfP  \bfLambda^{-1/2}\bfPsi_n)$ lie in the range $[26/27, 28/27]$ and consequently, $\tilde{\lambda}_j \geq 26 n\lambda_j/27$. \\

As mentioned earlier, the singular values of $K^{(\bfQ)}_{X_n, X_n}$ are the same as those of $\bfQ  \bfPsi_n \bfPsi_n^{\top}\bfQ  $. For the analysis, it suffices to have an upper bound on $\|K^{(\bfQ)}_{X_n, X_n}\|_2$, or equivalently, $\|\bfQ  \bfPsi_n \bfPsi_n^{\top}\bfQ  \|_2$. Using the result from~\citet[Proposition 3.8]{Moeller2021SamplingRKHS}, we know that the following relation holds with probability $1 - \delta/6$:
\begin{align*}
    \|\bfQ  \bfPsi_n \bfPsi_n^{\top}\bfQ  \|_2 \leq 2\left\{ 42 \log\left(\frac{12}{\delta}\right) T(R) , n\lambda_{R+1}  \right\}.
\end{align*}
Since $R \in \cR_{n, \tau}$, we can conclude that $\|K^{(\bfQ)}_{X_n, X_n}\|_2 = \|\bfQ  \bfPsi_n \bfPsi_n^{\top}\bfQ  \|_2 \leq 2\tau/27$. We are now ready to prove the lemma. \\

Using the relation $K_{X_n, X_n} = K^{(\bfP )}_{X_n, X_n} + K^{(\bfQ)}_{X_n, X_n}$, we can decompose the information gain of $X_n$ as follows:
\begin{align*}
    \tilde{\gamma}_{X_n, \tau} & = \frac{1}{2} \log\left(\det(I_n + \tau^{-1}K_{X_n, X_n})\right) \\
    & = \frac{1}{2} \log\left(\det(I_n + \tau^{-1}K^{(\bfP )}_{X_n, X_n} + \tau^{-1}K^{(\bfQ)}_{X_n, X_n})\right) \\
    & = \frac{1}{2} \log\left(\det( (I_n + \tau^{-1}K^{(\bfQ)}_{X_n, X_n})( I_n + \tau^{-1}(I_n + \tau^{-1}K^{(\bfQ)}_{X_n, X_n})^{-1}K^{(\bfP )}_{X_n, X_n}))\right) \\
    & = \frac{1}{2} \underbrace{\log(\det(I + \tau^{-1}K^{(\bfQ)}_{X_n, X_n}))}_{:=G_1} + \frac{1}{2} \underbrace{\log(\det(I + \tau^{-1}(I + \tau^{-1}K^{(\bfQ)}_{X_n, X_n})^{-1}K^{(\bfP )}_{X_n, X_n}))}_{ := G_2}.
\end{align*}
This decomposition is similar to that derived in~\citet[App. A, Eqn. 8]{Vakili2020InfoGain} with the roles of $K^{(\bfP )}_{X_n, X_n}$ and $K^{(\bfQ)}_{X_n, X_n}$ interchanged. \\

We begin with $G_1$. Since $\|K^{(\bfQ)}_{X_n, X_n}\|_2 \leq 2\tau/27$, all eigenvalues of $\tau^{-1}K^{(\bfQ)}_{X_n, X_n}$ are less than $1$. Using the relation $\log(1 + x) \geq x/2$, which holds for all $x \in [0,1]$, we can lower bound $G_1$ as follows:
\begin{align*}
    G_1 = \log(\det(I + \tau^{-1}K^{(\bfQ)}_{X_n, X_n})) \geq \frac{1}{2\tau}\tr(K^{(\bfQ)}_{X_n, X_n}).
\end{align*}
Note $k^{(\bfQ)}(X_i, X_i)$ are i.i.d. random variables with $\E[k^{(\bfQ)}(X_i, X_i)] = \sum_{r = R+1}^{\infty} \lambda_r$ and $|k^{(\bfQ)}(X_i, X_i)| \leq T(R)$. We can thus use Hoeffding inequality to obtain the following bound on $\tr(K^{(\bfQ)}_{X_n, X_n})$ which holds with probability at least $1 - \delta/6$:
\begin{align*}
    G_1 & \geq \frac{1}{2\tau} \tr(K^{(O)}_{X_n, X_n}) \\
    & \geq \frac{1}{2\tau} \left[n \sum_{r = R+1}^{\infty} \lambda_r - T(R) \sqrt{n \log(12/\delta)}\right] \\
    & \geq \frac{n T(R)}{2\tau F^2} \left( 1- F^2 \sqrt{\frac{\log(12/\delta)}{n}} \right) \\
    & \geq \frac{13n T(R)}{27\tau F^2}
\end{align*}
In the third line, we used the fact that $T(R) \leq F^2 \sum_{r = R+1}^{\infty} \lambda_r$ since $\|\varphi_j\|_{\infty} \leq F$ for all $j \in \N$ (Assumption~\ref{ass:eigenfunction_bound}). The fourth line uses the condition that $n \geq \overline{N}$. \\

To bound $G_2$, first note that using the condition on the spectrum on $\tau^{-1}K^{(\bfQ)}_{X_n, X_n}$, we can conclude that all the eigenvalues of $(I + \tau^{-1}K^{(\bfQ)}_{X_n, X_n})$ lie in the range $[1,2]$. Moreover, note that the spectrum of $(I + \tau^{-1}K^{(\bfQ)}_{X_n, X_n})^{-1}K^{(\bfP )}_{X_n, X_n}$ is the same as that of $(I + \tau^{-1}K^{(\bfQ)}_{X_n, X_n})^{-1/2}K^{(\bfP )}_{X_n, X_n}(I + \tau^{-1}K^{(\bfQ)}_{X_n, X_n})^{-1/2}$. On using Ostrowski's Theorem~\cite{Ostrowski1959Theorem} along with range of eigenvalues of $(I + \tau^{-1}K^{(\bfQ)}_{X_n, X_n})$, we can conclude that 
\begin{align*}
    G_2 =  \log(\det(I + \tau^{-1}(I + \tau^{-1}K^{(\bfQ)}_{X_n, X_n})^{-1}K^{(\bfP )}_{X_n, X_n})) \geq \log(\det(I + (2\tau)^{-1}K^{(\bfP )}_{X_n, X_n})).
\end{align*}
Using the relation for the eigenvalues of $K^{(\bfP )}_{X_n, X_n}$ derived earlier, we can further $G_2$ as follows:
\begin{align*}
    G_2 & \geq \log(\det(I + (2\tau)^{-1}K^{(\bfP )}_{X_n, X_n})) \\
    & \geq \sum_{j = 1}^R \log(1 + (2\tau)^{-1}\tilde{\lambda}_j) \\
    & \geq \sum_{j = 1}^R \log\left(1 + \frac{13n \lambda_j}{27\tau}\right) \\
    & \geq \sum_{j = 1}^R \frac{13n \lambda_j}{13n \lambda_j + 27\tau} \\
    & \geq \frac{13n}{27 F^2}\sup_{x \in \cX} \sum_{j = 1}^R \frac{\lambda_j}{n\lambda_j + \tau} \varphi_j^2(x).
\end{align*}
In the fourth line, we used the relation $\log(1 + x) \geq \frac{x}{x + 1}$, which holds for all $x \geq 0$. \\

On combining the bounds for $G_1$ and $G_2$, we obtain,
\begin{align*}
    \tilde{\gamma}_{X_n, \tau} & = \frac{1}{2}(G_1 + G_2) \\
    & \geq \frac{13n T(R)}{54\tau F^2} + \frac{13n}{54F^2} \sup_{x \in \cX} \sum_{j = 1}^R \frac{\lambda_j}{n\lambda_j + \tau} \varphi_j^2(x) \\
    & \geq \frac{13n}{54F^2}  \left( \sup_{x \in\cX}  \sum_{j = 1}^R \frac{\lambda_j}{n \lambda_j + \tau} \varphi^2(x) + \frac{T(R)}{\tau} \right) \\
    & \geq \frac{13n}{54F^2}  \sup_{x \in\cX}  \left(\sum_{j = 1}^{R} \frac{\lambda_j}{n\lambda_j + \tau} \varphi_j^2(x) + \sum_{j = R + 1}^{\infty} \frac{\lambda_j}{n\lambda_j + \tau} \varphi_j^2(x)  \right) \\
    & \geq \frac{13n}{54F^2} \sup_{x \in\cX} \psi_{x}^{\top} \bfZ^{-1} \psi_x,
\end{align*}
as required. Since each of the bounds on $G_1$ and the eigenvalues of $K^{(\bfP )}_{X_n, X_n}$ and $K^{(\bfQ)}_{X_n, X_n}$, holds with probability at least $1 - \delta/6$, the overall bound holds with probability at least $1 - \delta/2$.

\section{Proof of Theorems~\ref{thm:noiseless_regret} and~\ref{thm:noisy_regret}}
\label{appendix:proof_of_regret_theorems}

The proof of both the theorems is based along the lines of the proof of the Batched Pure Exploration (BPE) algorithm~\cite{Li2021BatchedPureExp}. We first begin with a brief discussion about Assumption~\ref{assumption:f_level_set_regularity} and then move on to the proof.

\begin{definition}
    Let $\Gamma: \cX \to \cX'$ be a map between two sets $\cX, \cX' \subset \R^d$. We call $\Gamma$ to be a bi-Lipschitz map if the inverse map, $\Gamma^{-1}$, exists and the following relations hold for some $L, L' > 0$:
    \begin{align*}
        \|\Gamma(x) - \Gamma(y)\|_2 & \leq L \|x -y \|_2 \ \ \forall x, y \in \cX \\
        \|\Gamma^{-1}(x) - \Gamma^{-1}(y)\|_2 & \leq L' \|x -y \|_2 \ \ \forall x, y \in \cX'.
    \end{align*}
    We refer to $(L, L')$ the Lipschitz constant pair of $\Gamma$. We also define normalized Lipschitz constant pair of $\Gamma$ to be the pair $(\tilde{L}, \tilde{L}') = \left( L  \left(\frac{\mathrm{vol}(\cX)}{\mathrm{vol}(\cX')}\right)^{1/d}, L' \left(\frac{\mathrm{vol}(\cX')}{\mathrm{vol}(\cX)}\right)^{1/d} \right)$.
\end{definition}

The normalized Lipschitz constant pair quantifies solely the change due to structure and discounts for the change in size between $\cX$ and $\cX'$. The following is a restatement of Assumption~\ref{assumption:f_level_set_regularity}.

\begin{assumption}
    Let $\cL_{\eta} = \{x \in \cX | f(x) \geq \eta \}$ denote the level set of $f$ for $\eta \in [-B, B]$. Then,
    \begin{itemize}
        \item For all $\eta \in [-B,B]$, $\cL_{\eta}$ is a disjoint union of at most $M_f < \infty$ closed, path connected components. 
        \item For a given $\eta \in [-B,B]$, let $\cL_{\eta}^{i}$ denote the $i^{\text{th}}$ such connected component of $\cL_{\eta}$. We assume that there exists a bi-Lipschitzian map $\Gamma_{\eta, i} : \cX \to \cL_{\eta}^{i}$ with normalized Lipschitz constant pair $\tilde{L}_{\eta, i}, \tilde{L}_{\eta, i}' > 0$ for all $\eta, i$. Let $L_f = \sup_{\eta, i} \tilde{L}_{\eta, i}$ and $L_f' = \sup_{\eta, i} \tilde{L}_{\eta, i}'$. We assume that $L_f, L_f' < \infty$.
    \end{itemize}
\end{assumption}

Assumption~\ref{assumption:f_level_set_regularity} is an assumption on the regularity of the level sets of the function $f$. The term $M_f$ can be thought of as the number of local maximas of $f$ and hence finiteness of $M_f$ is a mild assumption on $f$ satisfied by functions encountered in practice. Moreover, the knowledge of $M_f$ is only required for analysis and not for the algorithm to run. The second condition on $f$ is to ensure that the these connected components are topologically regular enough and to avoid certain pathological cases. In particular, the existence of a bi-Lipschitzian map between two sets implies topological similarity between the two sets. Intuitively, this assumption ensures that the shape of the level-sets is not ``too arbitrary''. Note that such an assumption on the level sets of $f$ is relatively mild as the RKHS endows smoothness properties to the function $f$ which translate to a degree of topological regularity of level sets~\cite{Giovanni2011SardTheorem, Lee2010TopologicalManifolds}.

\subsection{Proof of Theorem~\ref{thm:noiseless_regret}}
\label{appendix:proof_of_theorem_noiseless}

At a high level, the bound on regret is obtained by first separately bounding the regret during every epoch $r$ and then summing it across all epochs. During any epoch $r$, since REDS chooses points uniformly at random from the current domain $\cX_r$, we simply bound the regret incurred at each point queried during this epoch by the worst case scenario, i.e., $\varsigma_{r} := f(x^*) - \inf_{x \in \cX_r} f(x)$. This leads to an upper bound of  $\varsigma_r N_r M_f$ on the regret incurred during epoch $r$, as there are at most $M_f$ connected components in each level set. Since poorly performing regions of the domain are eliminated as the algorithm proceeds, $\inf_{x \in \cX_r} f(x)$ gets closer to $f(x^*)$, reducing the regret in each epoch as the algorithm proceeds.  \\

The following two lemmas ensure the correctness of the algorithm and help bound the regret incurred during each epoch.
\begin{lemma} \label{lemma:x_star_in_all_epochs}
    $x^* \in \cX_r$ for all $r \geq 1$.
\end{lemma}
\begin{lemma} \label{lemma:epoch_regret_noiseless}
    For all epochs $r$, we have,
    \begin{align*}
        \varsigma_r \leq \begin{cases} 2B & \text{ if } r = 1, \\ 4B \sup_{x \in \cX_{r-1}} \sigma_{r-1}(x) & \text{ if } r \geq 2. \end{cases}
    \end{align*}
\end{lemma}
We defer the proof of these lemmas to Appendix~\ref{sub:proof_auxiliary_lemma_regret}. Equipped with these lemmas, we move on to the proof of Theorem~\ref{thm:noiseless_regret}. The regret incurred by REDS can be bounded as
\begin{align*}
    R(T) & = \sum_{t = 1}^T f(x^*) - f(x_t) \leq \sum_{r = 1}^{S} \varsigma_r N_r M_f  \\
    & \leq 2B N_1 + 4B M_f \sum_{r = 2}^{S} \left[ N_r \cdot \sup_{x \in \cX_{r-1}} \sigma_{r-1}(x) \right].
\end{align*}
In the above expression, $S$ denotes the total number of epochs that begin during a run of REDS algorithm before reaching a total of $T$ queries. Since the epoch lengths double every epoch, we have $S \leq 1 + \log_{2}(T/N_1)$. We can further bound $R(T)$ using Lemma~\ref{lemma:variance_across_domains} (which in turn is based on Theorem~\ref{thm:concentration_random_sampling}) to bound the worst-case posterior standard deviation in the above equation. Since $\cX_{r-1}$ is compact ($\cX_{r-1}$ is closed by definition and $\cX_{r-1}$ is bounded because $\cX_{r-1} \subseteq \cX$) and $N_{r-1} \geq N_1 \geq C_{L_f, L_f'} \overline{N}$, we can invoke Lemma~\ref{lemma:variance_across_domains} to conclude 
\begin{align}
    R(T) & \leq 2B N_1 + 4B C_2 C_{L_f, L_f'}' M_f \sum_{r = 2}^{S} N_r \cdot  N_{r-1}^{(1 - \beta)/2} (\log (n/\delta'))^{\beta/2}, \label{eqn:noiseless_regret_in_beta_step_1}
\end{align}
where $\delta' = \delta/\log_2 T$, $C_2 = \sqrt{C_1}$ and $C_{L_f, L_f'}, C_{L_f, L_f'}'$ are the constants from Lemma~\ref{lemma:variance_across_domains} and depend only on $L_f, L_f'$. For simplicity, we define $C_f :=  C_{L_f, L_f'}' M_f$, as a constant that depends only on the function $f$. On plugging in the values of $N_r$, Eqn.~\eqref{eqn:noiseless_regret_in_beta_step_1} simplifies to 
\begin{align}
    R(T) & \leq 2B N_1 + 4B C_2 C_f \sum_{r = 2}^{S} N_r \cdot  N_{r-1}^{(1 - \beta)/2} (\log (n/\delta'))^{\beta/2} \nonumber \\
    & \leq 2B N_1 + 4BC_2 C_f  N_1^{(3 - \beta) /2}\sum_{r = 2}^{S} 2^{r-1} \cdot  2^{(r-2)(1 - \beta)/2} \left(\log \left( \frac{N_{1}}{\delta'} \cdot 2^{r-2}\right)\right)^{\beta/2} \nonumber \\
    & \leq 2B N_1 + 8BC_2 C_f  N_1^{(3 - \beta) /2}\sum_{r = 0}^{S -2 } 2^{r(3 - \beta)/2} \left(\log \left( \frac{N_{1}}{\delta'} \cdot 2^{r}\right)\right)^{\beta/2}. \label{eqn:noiseless_regret_in_beta_step_2}
\end{align}

We consider three separate cases based on the value of $\beta$:
\begin{itemize}
    \item $\beta < 3$: Under this case, Eqn.~\eqref{eqn:noiseless_regret_in_beta_step_2} can be simplified as follows:
    \begin{align*}
        R(T) & \leq 2B N_1 + 8BC_2 C_f  N_1^{(3 - \beta) /2}\sum_{r = 0}^{S -2 } 2^{r(3 - \beta)/2} \left(\log \left( \frac{N_{1}}{\delta'} \cdot 2^{r}\right)\right)^{\beta/2} \\
        & \leq 2B N_1 + 8BC_2 C_f  N_1^{(3 - \beta) /2} \left(\log \left( \frac{T}{\delta'}\right)\right)^{3/2} \sum_{r = 0}^{S -2 } 2^{r(3 - \beta)/2}  \\
        & \leq 2B N_1 + 8BC_2 C_f  N_1^{(3 - \beta) /2} \left(\log \left( \frac{T}{\delta'}\right)\right)^{3/2}  \frac{2^{(S - 1)(3 - \beta)/2} - 1}{2^{(3 - \beta)/2} - 1}  \\
        & \leq 2B N_1 +  \frac{8BC_2 C_f}{2^{(3 - \beta)/2} - 1} T^{(3 - \beta) /2} \left(\log \left( \frac{T}{\delta'}\right)\right)^{3/2}.
    \end{align*}
    \item $\beta = 3$: For this value of $\beta$, Eqn.~\eqref{eqn:noiseless_regret_in_beta_step_2} can be simplified as follows:
    \begin{align*}
        R(T) & \leq 2B N_1 + 8BC_2 C_f  N_1^{(3 - \beta) /2}\sum_{r = 0}^{S -2 } 2^{r(3 - \beta)/2} \left(\log \left( \frac{N_{1}}{\delta'} \cdot 2^{r}\right)\right)^{\beta/2} \\
        & \leq 2B N_1 + 8BC_2C_f  \cdot \left(\log \left( \frac{T}{\delta'}\right)\right)^{3/2} \cdot \sum_{r = 0}^{S -2 } 1  \\
        & \leq 2B N_1 + 8BC_2 C_f  \cdot \left(\log \left( \frac{T}{\delta'}\right)\right)^{3/2} \cdot \log \left( \frac{T}{N_1} \right).
    \end{align*}
    \item $\beta > 3$: For this range, we have,
    \begin{align*}
        R(T) & \leq 2B N_1 + 8BC_2 C_f  N_1^{(3 - \beta) /2}\sum_{r = 0}^{S -2 } 2^{r(3 - \beta)/2} \left(\log \left( \frac{N_{1}}{\delta'} \cdot 2^{r}\right)\right)^{\beta/2} \\
        & \leq 2B N_1 + 8BC_2 C_f  \cdot \left(\log \left( \frac{T}{\delta'}\right)\right)^{3/2} \cdot \sum_{r = 0}^{S -2 } 2^{r(3 - \beta)/4} \left[ \frac{\log(N_1 \cdot 2^r) + \log(1/\delta')}{N_1 \cdot 2^{r/2}} \right]^{(\beta -3)/2} \\
        & \leq 2B N_1 + 8BC_2C_f  \cdot \left(\log \left( \frac{T}{\delta'}\right)\right)^{3/2} \cdot \left[ \frac{\log(N_1/\delta')}{N_1} \right]^{(\beta -3)/2} \cdot \sum_{r = 0}^{S -2 } 2^{r(3 - \beta)/4}  \\
        & \leq 2B N_1 + 8BC_2 C_f  \cdot \left(\log \left( \frac{T}{\delta'}\right)\right)^{3/2} \cdot \left[ \frac{\log(N_1/\delta')}{N_1} \right]^{(\beta -3)/2} \cdot \sum_{r = 0}^{\infty} 2^{r(3 - \beta)/4}  \\
        & \leq 2B N_1 + \frac{8BC_2 C_f}{1 - 2^{(3 - \beta)/4}} \cdot \left(\log \left( \frac{T}{\delta'}\right)\right)^{3/2}.
    \end{align*}
    In the third step, we used the fact that $\dfrac{\log(N_1 \cdot 2^r/\delta')}{N_1 \cdot 2^{r/2}}$ is a decreasing function of $r$ for all $r \geq 0$ and in the fifth step we used the fact that $N_1 \geq \log(N_1/\delta')$ since $N_1 \geq \overline{N}(\delta')$.
\end{itemize}

On combining all the cases, we arrive at the result. The statement in Corollary~\ref{corollary:noise_free} follows immediately from the above proof by plugging in $\beta = 1 + 2\nu/d$.

\subsection{Proof of Theorem~\ref{thm:noisy_regret}}
\label{appendix:proof_of_theorem_noisy}

The proof of Theorem~\ref{thm:noisy_regret} is almost identical to that of Theorem~\ref{thm:noiseless_regret}. The following lemma is a counterpart to Lemma~\ref{lemma:epoch_regret_noiseless} for the noisy case.
\begin{lemma} \label{lemma:epoch_regret_noisy}
    For all epochs $r$, the following relation holds with probability at least $1 - \delta/2$:
    \begin{align*}
        \varsigma_r \leq \begin{cases} 2B & \text{ if } r = 1, \\ 4\alpha_{\tau}(\delta'/2) \left[\sup_{x \in \cX_{r-1}} \sigma_{r-1, \tau}(x)\right]  + \frac{2B}{T} + R \sqrt{\frac{2}{T\tau}\log\left(\frac{4T}{\delta'} \right)} & \text{ if } r \geq 2. \end{cases}
    \end{align*}
\end{lemma}
The proof of this lemma is identical to that of Lemma~\ref{lemma:epoch_regret_noiseless} with the definitions of $\UCB$ and $\LCB$ changed according to the noisy setup (See~\cite{Vakili2021OptimalSimpleRegret} for an exact derivation). On using Lemma~\ref{lemma:variance_across_domains} (for the noisy case) along with Lemma~\ref{lemma:epoch_regret_noisy}, we can rewrite Eqn.~\eqref{eqn:noiseless_regret_in_beta_step_1} as
\begin{align}
    R(T) & \leq 2B N_1 +  M_f \sum_{r = 2}^{S} N_r \cdot  \left[ 4 \sqrt{C_{\tau}} C_{L_f, L_f'}' \alpha_{\tau}(\delta'/2) \sqrt{\frac{\gamma_{N_{r-1}, \tau}}{N_{r-1}}} + \frac{2B}{T} + R \sqrt{\frac{2}{T\tau}\log\left(\frac{4T}{\delta'} \right)}\right] \nonumber \\
    & \leq 2B N_1 +  M_f \sum_{r = 2}^{S} N_r \cdot  \left[ 4 \sqrt{C_{\tau}} C_{L_f, L_f'}' \alpha_{\tau}(\delta'/2) \sqrt{\frac{\gamma_{T, \tau}}{N_{r-1}}} + \frac{2B}{T} + R \sqrt{\frac{2}{T\tau}\log\left(\frac{4T}{\delta'} \right)}\right], \label{eqn:noisy_regret_step_1}
\end{align}
where second line follows using monotonicity of $\gamma_{n, \tau}$ i.e., $\gamma_{n_1, \tau} \leq \gamma_{n_2, \tau}$ for all $n_1 \leq n_2$ and $C_{\tau}$ is the leading constant in Eqn.~\eqref{eqn:sigma_noisy_bound}. On plugging in the values of $N_r$ in Eqn.~\eqref{eqn:noisy_regret_step_1}, we obtain,
\begin{align*}
    R(T) & \leq 2B N_1 +  \sum_{r = 2}^{S} N_r \cdot  \left[ 4 \sqrt{C_{\tau}} C_{L_f, L_f'}' M_f \alpha_{\tau}(\delta'/2) \sqrt{\frac{\gamma_{T, \tau}}{N_{r-1}}} + \frac{2BM_f}{T} +  R M_f \sqrt{\frac{2}{T\tau}\log\left(\frac{4T}{\delta'} \right)}\right] \\
    & \leq 2B N_1 +  \sum_{r = 2}^{S}  \left[ 4 \sqrt{N_1 C_{\tau}} C_f \alpha_{\tau}(\delta'/2) \sqrt{\gamma_{T, \tau}} \cdot 2^{r-1} \cdot 2^{-(r-2)/2} +  M_f \cdot \frac{2BN_1}{T} \cdot 2^{r-1} + M_f \cdot RN_1  \sqrt{\frac{2}{T\tau}\log\left(\frac{4T}{\delta'}  \right)} \cdot 2^{r-1} \right] \\
    & \leq 2B N_1 +    8 \sqrt{N_1 C_{\tau}} C_f \alpha_{\tau}(\delta'/2) \sqrt{\gamma_{T, \tau}} \left( \sum_{r = 0}^{S-2} 2^{r/2} \right) + M_f \cdot \left( \frac{4B}{T} + 2R \sqrt{\frac{2}{T\tau}\log\left(\frac{4T}{\delta'} \right)} \right) N_1  \left( \sum_{r = 0}^{S-2} 2^{r} \right) \\
    & \leq 2B N_1 + \frac{8}{\sqrt{2} - 1} \sqrt{N_1 C_{\tau}} C_f \alpha_{\tau}(\delta'/2) \sqrt{\gamma_{T, \tau}} \cdot \sqrt{\frac{T}{N_1}} + M_f \cdot \left( \frac{4B}{T} + 2R \sqrt{\frac{2}{T\tau}\log\left(\frac{4T}{\delta'} \right)} \right)  \cdot N_1    \cdot \frac{T}{N_1} \\
    & \leq 2B N_1 + \frac{8}{\sqrt{2} - 1} \sqrt{C_{\tau}} C_f \alpha_{\tau}(\delta'/2) \sqrt{T\gamma_{T, \tau}}  + 4BM_f + 2R M_f \sqrt{\frac{2T}{\tau}\log\left(\frac{4T}{\delta'} \right)}, 
\end{align*}
where $C_f = C_{L_f, L_f'}' M_f$ as before. Hence, $R(T)$ satisfies $\tilde{\cO}(\sqrt{T\gamma_{T, \tau}})$, as required.

\subsection{Proof of Auxiliary Lemmas}
\label{sub:proof_auxiliary_lemma_regret}

\subsubsection{Proof of Lemma~\ref{lemma:x_star_in_all_epochs}}

The main ingredient in the proof is the relation: $|f(x) - \mu_{r-1}(x)| \leq B \sigma_{r-1}(x)$, which holds for all $x \in \cX_{r-1}$ and across all epochs $r$. This is a well-known relation in the literature~\cite{Vakili2021OptimalSimpleRegret, Lyu2020efficient} that bounds the predictive performance of the posterior mean in terms of posterior variance. \\

We use induction to prove the lemma. Since $\cX_1 = \cX$ and $x^* \in \cX$ holds by definition, $x^* \in \cX_1$. Assume that $x^* \in \cX_{r-1}$.  Using the relation $|f(x) - \mu_{r-1}(x)| \leq B \sigma_{r-1}(x)$, we can conclude,
\begin{align*}
    \sup_{x' \in \cX_{r-1}} \LCB_{r-1}(x') = \sup_{x' \in \cX_{r-1}} (\mu_{r-1}(x') - B\sigma_{r-1}(x')) \leq \sup_{x' \in \cX_{r-1}} f(x') = f(x^*) \leq \UCB_{r-1}(x^*),
\end{align*}
where we used the inductive hypothesis to establish $\sup_{x' \in \cX_{r-1}} f(x') = f(x^*)$. This implies that $x^* \in \cX_r$, as required. \\

\subsubsection{Proof of Lemma~\ref{lemma:epoch_regret_noiseless}}

We separately show the bounds for $r = 1$ and $r \geq 2$. For the first epoch, we have, 
\begin{align*}
    \varsigma_{1} = f(x^*) - \inf_{x \in \cX_1} f(x) = f(x^*) - \inf_{x \in \cX} f(x) \leq 2 \sup_{x \in \cX} f(x) \leq 2 B.
\end{align*}
We used the fact that $\sup_{x \in \cX} f(x) = \sup_{x \in \cX} f^{\top} \psi_x \leq \sup_{x \in \cX} \|f\|_{\cH_k} \|\psi_x\|_{\cH_k} \leq B$. Consider any epoch $r \geq 2$. For the analysis, we define 
\begin{align*}
    \cX_r' := \{x \in \cX_{r-1} : f(x) + 2B\sigma_{r-1}(x) \geq \sup_{x' \in \cX_{r-1}} f(x') - 2B\sigma_{r-1}(x') \}.
\end{align*}
The region $\cX_r'$ satisfies $\cX_r \subseteq \cX_r'$. To establish this, we once again employ the relation $|f(x) - \mu_{r-1}(x)| \leq B \sigma_{r-1}(x)$. Using the relation, we can conclude that
\begin{align*}
    \UCB_{r-1}(x) & = \mu_{r-1}(x) + B \sigma_{r-1}(x) \leq  (f(x) + B \sigma_{r-1}(x)) + B \sigma_{r-1}(x) = f(x) + 2B\sigma_{r-1}(x) \\
    \LCB_{r-1}(x) & = \mu_{r-1}(x) - B \sigma_{r-1}(x) \geq  (f(x) - B \sigma_{r-1}(x)) - B \sigma_{r-1}(x) = f(x) - 2B\sigma_{r-1}(x).
\end{align*}
The inclusion $\cX_r \subseteq \cX_r'$ follows immediately from the definition of $\cX_r$ and $\cX_r'$ and the above expressions. \\

Consider the following relation which holds for any $x \in \cX_r'$.
\begin{align}
    f(x) + 2B\sigma_{r-1}(x) & \geq \sup_{x' \in \cX_{r-1}} f(x') - 2B\sigma_{r-1}(x') \nonumber \\
    \implies f(x) & \geq \sup_{x' \in \cX_{r-1}} [f(x') - 2B\sigma_{r-1}(x')] - \sup_{x'' \in \cX_{r-1}} [ 2B\sigma_{r-1}(x'')]  \nonumber \\
    & \geq \sup_{x' \in \cX_{r-1}} f(x') - \sup_{x'' \in \cX_{r-1}} [ 4B\sigma_{r-1}(x'')] \nonumber \\
    & \geq f(x^*) - \sup_{x'' \in \cX_{r-1}} [ 4B\sigma_{r-1}(x'')]. \label{eqn:bound_on_varsigma_analysis}
\end{align}
In the last line, we used Lemma~\ref{lemma:x_star_in_all_epochs} to conclude $\sup_{x' \in \cX_{r-1}} f(x') = f(x^*)$. Since $\cX_r \subset \cX_r'$, we can use Eqn.~\eqref{eqn:bound_on_varsigma_analysis} to obtain an upper bound on $\varsigma_r$ as follows:
\begin{align*}
    \varsigma_r & = f(x^*) - \inf_{x \in \cX_{r}} f(x) \\
    & \leq f(x^*) - \inf_{x \in \cX_{r}'} f(x) \\
    & \leq f(x^*) - \left[ f(x^*) - \sup_{x' \in \cX_{r-1}}  4B\sigma_{r-1}(x') \right] \\
    & \leq 4B \sup_{x' \in \cX_{r-1}}  \sigma_{r-1}(x').
\end{align*}

\subsubsection{Proof of Lemma~\ref{lemma:variance_across_domains}}

We begin with the noiseless case. For brevity, we drop the subscript $0$ from the posterior variance corresponding to the noiseless case. Consider a kernel $k$ and let $\cH$ and $\cH'$ denote the RKHS induced by $k$ on $\cX$ and $\cX'$. Since $\cX' \subset \cX$, it is straightforward to note that $\cH' \subseteq \cH$. Using the result from~\citet[Theorem 10.46]{Wendland2004}, we know that for every $f \in \cH'$ there exists a natural extension $\mathscr{E}f \in \cH$ such that $\|\mathscr{E}f\|_{\cH} = \|f\|_{\cH'}$. Consequently, we can conclude $\{ f : \|f\|_{\cH'} \leq 1 \} \subseteq \{ f: \|f\|_{\cH} \leq 1 \}$. Lastly, note that $\cH'$ is same as the RKHS of the kernel $k'(x,y) = k(\Gamma(x), \Gamma(y))$ over the domain $\cX$. Here $\Gamma$ denotes the bi-Lipschitian map $\Gamma: \cX \to \cX'$ as given by Assumption~\ref{assumption:f_level_set_regularity}. \\

Let $X \subset \cX$ be any set of distinct points and $\sigma_{X}'(x)$ and  $\sigma_{X}(x)$ denote the posterior standard deviation at any point $x$ computed using the kernels $k'$ and $k$. Using the dual formulation of posterior variance, we have the following relation:
\begin{align*}
    \sigma_{X}'(x) = \sup_{\substack{f \in \cH' \\ \|f\|_{\cH'} \leq 1 \\ f(X) = \{0\}}} f(x) \leq  \sup_{\substack{f \in \cH \\ \|f\|_{\cH} \leq 1 \\ f(X) = \{0\}}} f(x) = \sigma_{X}(x).
\end{align*}
In the above relation, we used the fact that $\cH' \subset \cH$ and the unit ball in $\cH'$ is contained in the unit ball in $\cH$. This implies that the prediction made using the kernel $k'$ has a smaller error than the prediction made by using kernel $k$. If we set $X = \Gamma^{-1}(X')$\footnote{For any operator $\Gamma$ and $X = \{x_1, x_2, \dots, x_n\}$, we use the shorthand $\Gamma(X)$ for the set $\{\Gamma(x_1), \Gamma(x_2), \dots, \Gamma(x_n)\}$.}, then the above is equivalent to saying that the prediction error using kernel $k$ corresponding to set of points $X' \in \cX'$ is smaller than the prediction error using kernel $k$ corresponding to set of points $X \in \cX$.   \\

Since the points $X'$ are distributed uniformly in $\cX'$, the points $X = \Gamma^{-1}(X')$ are distributed according to density $\vartheta(x) = \frac{\det(\nabla\Gamma(x))}{\mathrm{vol}(\cX')} $ for all $x \in \cX$, where $\det(A)$ denotes the determinant of a matrix $A$ and $\nabla\Gamma$ denotes the Jacobian of $\Gamma$. Note that $\nabla\Gamma$ (and hence the density $\vartheta$) is well-defined almost everywhere (a.e.) as a consequence of Rademacher's theorem~\citep[Chp. 7]{Rudin1987Analysis} and Lipschitz continuity of $\Gamma$. \\

Let $\varrho_{\mathrm{unif}}$ denote the uniform distribution on $\cX$ (i.e., the Lebesgue measure). We construct a (random) subset of $X$, denoted by $Y$, as follows. Each point $x_i$ for $i \in \{1,2,\dots, n\}$ is added into $Y$ independently of others with probability $c_{\vartheta} \frac{\varrho_{\mathrm{unif}}(x_i)}{\vartheta(x_i)}$, where $c_{\vartheta} = \inf_{x} \frac{\vartheta(x)}{\varrho_{\mathrm{unif}}(x)}$ (where the infimum is taken over where $\vartheta$ is well defined). It is straightforward to note that the samples in $Y$ are distributed according to $\varrho_{\mathrm{unif}}$. Using the Bernstein inequality for sum of Bernoulli random variables, we can conclude that $|Y|$, the number of points in $Y$ satisfies the relation $|Y| \geq \frac{c_{\vartheta}n}{2C_{\vartheta}}$ with probability $1 - \delta$ as long as $\frac{3c_{\vartheta}n}{16C_{\vartheta}} \geq \log(2/\delta)$. Here $C_{\vartheta} = \sup_{x} \frac{\vartheta(x)}{\varrho_{\mathrm{unif}}(x)}$. Since $Y \subseteq X$, the prediction based on the values of $X$ is no worse than the prediction based on the values of $Y$. Thus,
\begin{align*}
    \sup_{x' \in \cX'} \sigma_{X'}^2(x') & \leq \sup_{x \in \cX} \sigma_{X}^2(x)  \leq \sup_{x \in \cX} \sigma_{Y}^2(x)  
\end{align*}
An identical result holds for the noisy case using an identical series of arguments using the kernel $k_{\tau}(x, x') = k(x, x) + \tau \delta_{x = x'}$~\cite{Kanagawa2018}, where $\delta_{x = x'}$ denotes the dirac delta function. We can invoke the result from Theorem~\ref{thm:concentration_random_sampling} for uniform samples on $\cX$ to bound $\sigma_{Y}^2(x)$ under both the noisy and noiseless settings to obtain the following relations
\begin{align*}
    \sup_{x' \in \cX'} \sigma_{X', \tau}^2(x') \leq \sup_{x \in \cX} \sigma_{Y, \tau}^2(x)  \leq \frac{C_{\vartheta}}{c_{\vartheta}} \cdot \frac{216}{13} \cdot F^2 \tau  \cdot \frac{\gamma_{n, \tau}}{n}, \\
    \sup_{x' \in \cX'} \sigma_{X', 0}^2(x') \leq \sup_{x \in \cX} \sigma_{Y, 0}^2(x)  \leq \frac{C_{\vartheta}}{c_{\vartheta}} \cdot \frac{216}{13} \cdot F^2  \cdot n^{1- \beta}.
\end{align*}

We only need to obtain a bound the ratio $C_{\vartheta}/c_{\vartheta}$ that is independent of $n$ to complete the proof. Using the Lipschitzness of $\Gamma$ and $\Gamma^{-1}$, we can conclude that 
\begin{align*}
    L_f'^{-d} \leq |\det(\nabla \Gamma)| \leq L_f^{d}.
\end{align*}
Using the definition of $c_{\vartheta}$, we have,
\begin{align*}
    c_{\vartheta} = \inf_{x} \frac{\vartheta(x)}{\varrho_{\mathrm{unif}}(x)} = \inf_{x} \frac{\det(\nabla\Gamma(x)) \mathrm{vol}(\cX)}{\mathrm{vol}(\cX')} \geq \frac{\mathrm{vol}(\cX)}{L_f'^d\mathrm{vol}(\cX')} = \tilde{L_f}'^{-d}.
\end{align*}
Similarly, 
\begin{align*}
    C_{\vartheta} = \sup_{x} \frac{\vartheta(x)}{\varrho_{\mathrm{unif}}(x)} = \sup_{x} \frac{\det(\nabla\Gamma(x)) \mathrm{vol}(\cX)}{\mathrm{vol}(\cX')} \leq \frac{L_f^d\mathrm{vol}(\cX)}{\mathrm{vol}(\cX')} = \tilde{L_f}^{d}.
\end{align*}
Hence, $C_{\vartheta}/c_{\vartheta} \leq (\tilde{L}_f/\tilde{L}_f')^d$ depends only on $(\tilde{L}_f, \tilde{L}_f')$ and is independent of $n$, as required.

\end{document}